\documentclass[fleqn,10pt]{wlscirep}
\usepackage[utf8]{inputenc}
\usepackage[T1]{fontenc}
\usepackage[table,dvipsnames]{xcolor}  

\usepackage{amsmath}
\usepackage{array}
\usepackage{longtable}
\usepackage{geometry}
\usepackage{ragged2e}
\usepackage{booktabs}
\usepackage{tabularx}
\usepackage{multicol}
\usepackage{makecell}
\usepackage{pdflscape}
\usepackage{graphicx}
\usepackage{subcaption}
\usepackage{adjustbox}

\newcommand{\redbar}[1][70]{\textcolor{red!#1!}{\rule{4ex}{2ex}}}
\newcommand{\bluebar}[1][70]{\textcolor{blue!#1!}{\rule{4ex}{2ex}}}
\newcommand{\violetbar}[1][70]{\textcolor{violet!#1!}{\rule{4ex}{2ex}}}
\newcommand{\orangebar}[1][70]{\textcolor{orange!#1!}{\rule{4ex}{2ex}}}

\newcolumntype{P}[1]{>{\RaggedRight\arraybackslash}p{#1}}

\title{CodeCytos: AI-assisted spatial molecular imaging analysis via code-augmented agent action space}

\author[1,\dag]{Hung Q. Vo}
\author[1,\dag]{Huy Q. Vo}
\author[1]{Son T. Ly}
\author[2]{Zhihao Wan}
\author[1]{Anh-Vu Nguyen}
\author[2]{Hong Zhao}
\author[2]{Jianting Sheng}
\author[2,\ddag]{Stephen T. C. Wong}
\author[1,\ddag,\S]{Hien V. Nguyen}

\affil[1]{University of Houston, Department of Electrical and Computer Engineering, Houston, Texas, 77004, USA}
\affil[2]{Houston Methodist Hospital, Department of Systems Medicine and Biomedical Engineering, Houston, Texas, 77030, USA}

\affil[ ]{$^{\dag}$These authors are co-first authors.}
\affil[ ]{$^{\ddag}$These authors are co-senior authors.}
\affil[ ]{$^{\S}$Correspondence: Hien V. Nguyen (hvnguy35@central.uh.edu)}

\begin{abstract}
    Conventional tissue image analysis software provides foundational capabilities for cellular analysis, including segmentation, basic morphological feature extraction, and spatial organization analysis. However, these tools often require manual intervention and are not well integrated with code-driven automation, limiting efficiency and scalability for complex spatial tissue studies. In addition, they offer limited flexibility for custom analyses, as they typically support only a fixed set of pre-implemented spatial cellular features. To address these limitations, we propose CodeCytos, a coding-based reasoning agent framework that enables dynamic, programmable interaction with spatial molecular imaging data, to improve automation and customization. CodeCytos is designed to streamline the exploration of custom spatial cellular features and adapt to diverse research needs. We demonstrate its utility through case studies on four expert-curated datasets from distinct tissue types: frontal cortex, non-small-cell lung cancer, pancreas, and tonsil. We evaluate CodeCytos under a realistic minimal prompt setting, where bioscientists pose simple questions without task-specific instructions or contextual information about spatial cellular analysis, and benchmark multiple LLM backbones with strong coding capabilities. We further show that incorporating tailored, domain-agnostic few-shot in-context coding–reasoning examples (randomly sampled demonstrations outside the spatial analysis domain) can substantially improve performance without requiring costly, expert-crafted in-domain demonstrations. Overall, CodeCytos outperforms baseline approaches, highlighting the potential of code-action agents to assist with custom feature exploration in spatial molecular imaging and to accelerate biomarker discovery.
\end{abstract}
\begin{document}

\flushbottom
\maketitle

%
%


\section*{Introduction}

Spatial molecular image analysis is crucial for understanding complex tissue microenvironments and generating insights that support biomarker discovery. However, extracting meaningful features from cellular tissue images often requires specialized expertise in image analysis as well as artificial intelligence and machine learning. Existing analysis software—such as QuPath \cite{bankhead2017qupath}, HALO \cite{oxford_halo_thl}, Visiopharm \cite{visiopharm_website}, Napari \cite{chiu2022napari}, FIJI/ImageJ \cite{schindelin2015imagej}, Squidpy \cite{palla2022squidpy}, and CellProfiler \cite{carpenter2006cellprofiler}—typically provides only a limited set of predefined functions for morphological feature extraction and spatial organization analysis. Consequently, biomedical researchers may be limited in their ability to formulate and evaluate custom hypotheses, restricting the full use of these data for biomarker exploration.

\begin{figure}
    \centering

    \begin{subfigure}{\linewidth}
        \centering
        \rlap{\raisebox{\dimexpr\height-0.8\baselineskip\relax}{\hspace{0.8\baselineskip}\textbf{a}}}%
        \includegraphics[width=\linewidth]{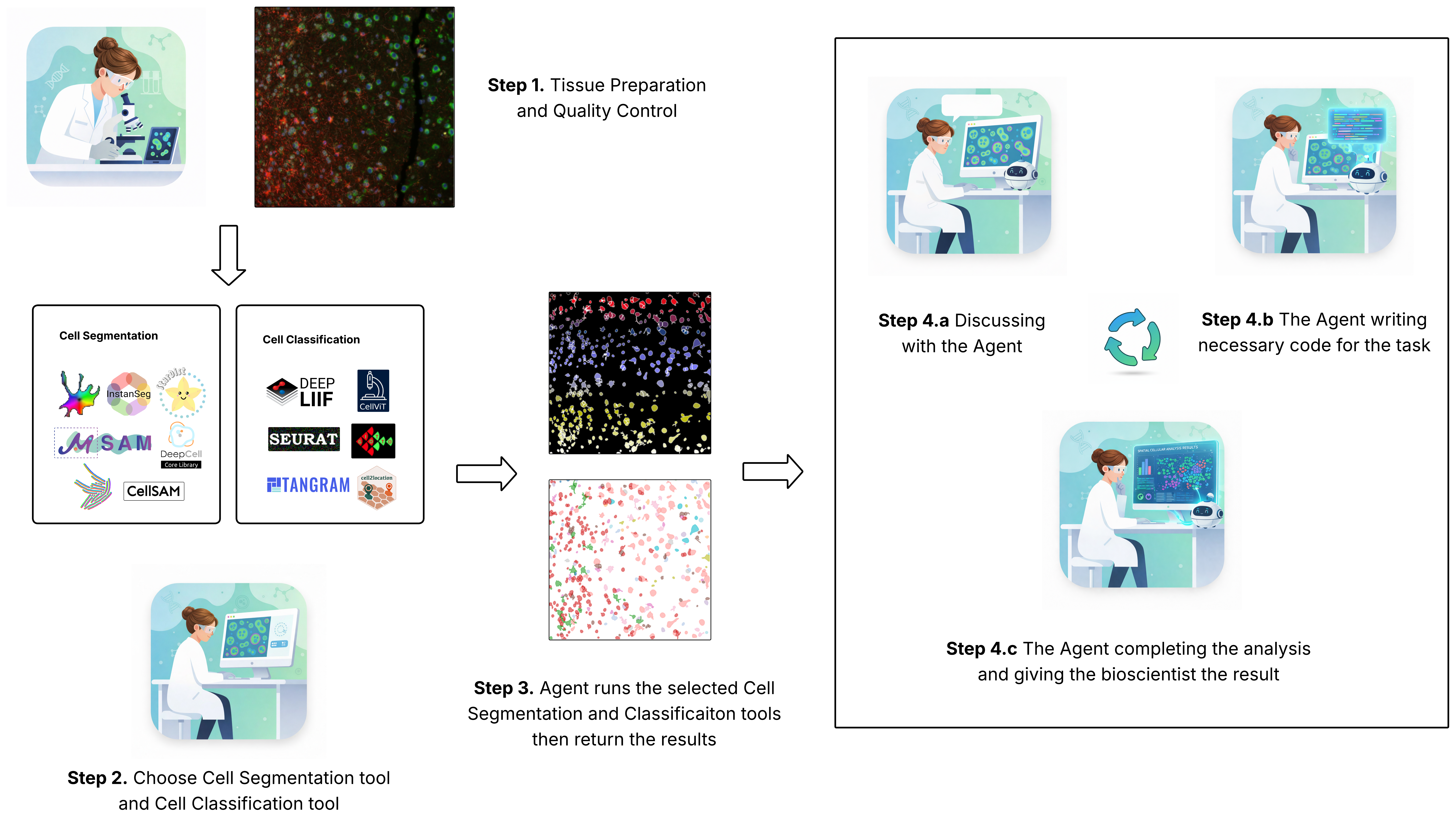}
        \label{fig:CodeCytos_workflow}
    \end{subfigure}

    \vspace{0.5cm}

    \begin{minipage}{\linewidth}
        \centering
        \begin{subfigure}{\textwidth}
            \centering
            \rlap{\raisebox{\dimexpr\height-0.8\baselineskip\relax}{\hspace{0.8\baselineskip}\textbf{b}}}%
            \includegraphics[width=\linewidth]{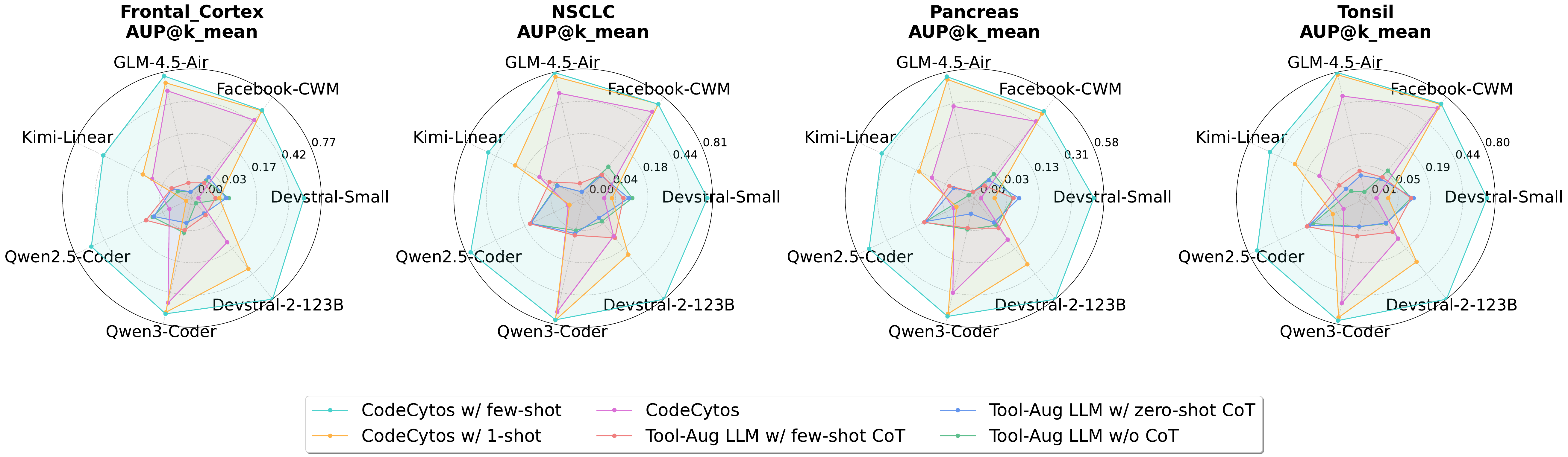}
            \label{fig:CodeCytos_result}
        \end{subfigure}
    \end{minipage}

    \caption{\textbf{CodeCytos Agent Usage Workflow and Benchmarking Performance.} \textbf{a,} Proposed workflow for bioscientists using CodeCytos: Bioscientists first prepare tissue samples and upload the digitized data to the pipeline. Next, they select appropriate cell-segmentation and cell-classification tools, consulting AI scientists as needed. CodeCytos then configures and applies the chosen segmentation and classification methods. Bioscientists can subsequently request specific spatial features via a text query. The CodeCytos agent performs the required reasoning, code generation, and execution, and returns the results upon completion. \textbf{b,} Benchmark results reported as the mean area under the Pass@k curve (AUP@k) across all questions in the Spatial Molecular Imaging Benchmark, spanning four tissue types: Frontal Cortex, NSCLC, Pancreas, Tonsil. We use LLM with strong coding abilities: Qwen2.5-Coder-32B-Instruct, Qwen3-Coder-30B-Instruct, Kimi-Linear-48B-Instruct, GLM-4.5-Air, Devstral-2-123B-Instruct, Devstral-Small, Facebook-CWM. Two main settings are evaluated: 1) Tool-Augmented LLM and 2) Our proposed CodeAct-based CodeCytos. For each main setting, three experiments are further carried out. For Tool-Augmented LLM, three subsettings are 1.a) Without Chain-of-Thought, 1.b) Zero-shot Chain-of-Thought, and 1.c) Few-shot Chain-of-Thought. For CodeCytos, three subsettings are: 2.a) Original CodeAct (Zero-shot), 2.b) One-shot CodeAct, and 2.c) Few-shot CodeAct.}
    
    \label{fig:CodeCytos_agent}
\end{figure}

To address these limitations, we introduce CodeCytos (figure \ref{fig:CodeCytos_agent}), a coding-based reasoning LLM agent that assists biomedical scientists in extracting spatial cellular features from molecular tissue images in response to natural-language queries. CodeCytos is built on a code-action (CodeAct) agent framework \cite{wang2024executable}, enabling it to interpret a researcher’s requested spatial feature and execute the corresponding analysis. Given a text query, the agent first parses the request, next it need to reason (Thought) about the next step, generate and executes code (Code), and then observe (Observation) the environment (e.g., prior reasoning, generated code, execution outputs, logs, and raised errors). It uses these observations as context to determine subsequent actions. This Thought–Code–Observation cycle repeats until the requested spatial feature extraction is completed.

Several studies have explored agentic systems for diverse tasks across biomedical domains. For example, Zhou et al. \cite{zhou2024ai} developed a fully automated agent for multi-omics analysis. Wang et al. \cite{wang2025spatialagent} proposed an agent for spatial-omics analysis that can operate either autonomously or as a copilot. Adibvafa et al. \cite{fallahpourmedrax} presented an AI agent for chest X-ray analysis that integrates task-specific chest X-ray models with multimodal large language models in a unified ReAct-based framework; the system supports tool-driven workflows spanning classification, segmentation, visual question answering, and report generation. Similarly, WSI-Agents by Lyu et al. \cite{lyu2026wsi} proposed a multi-agent decision-support system for pathologists that combines the flexibility of multimodal LLMs with specialized whole-slide imaging models and pathology knowledge bases, enabling tasks such as morphology analysis, diagnosis, treatment planning, and report generation.

\begin{figure}
    \centering

    \begin{subfigure}{0.9\linewidth}
        \centering
        \rlap{\raisebox{\dimexpr\height-0.8\baselineskip\relax}{\hspace{0.8\baselineskip}\textbf{a}}}%
        \includegraphics[width=\linewidth]{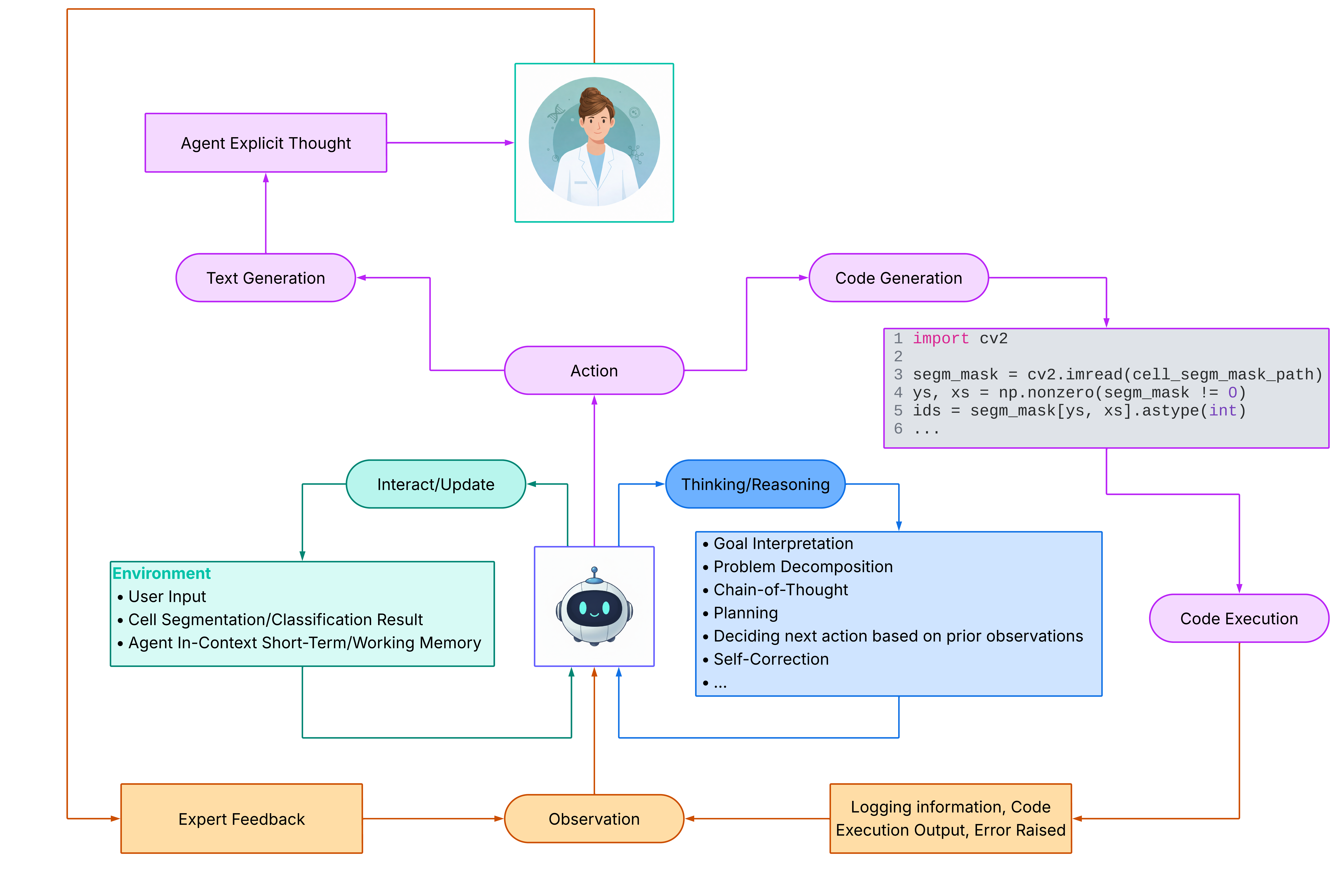}
        \label{fig:CodeCytos_environment}
    \end{subfigure}


    \begin{subfigure}{0.85\linewidth}
        \centering
        \rlap{\raisebox{\dimexpr\height-0.8\baselineskip\relax}{\hspace{0.8\baselineskip}\textbf{b}}}%
        \includegraphics[width=\linewidth]{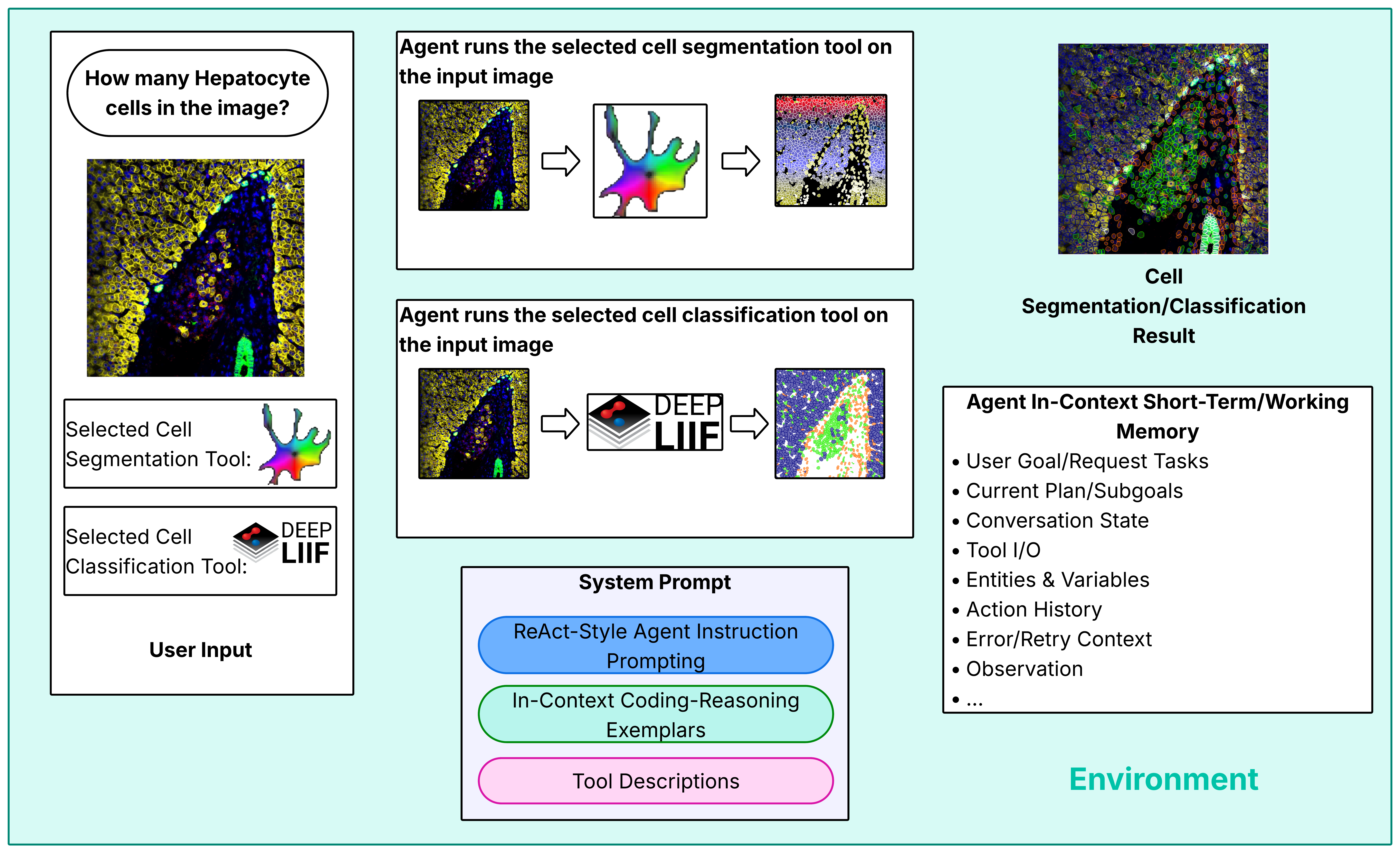}
    \end{subfigure}
    \caption{\textbf{Architecture Diagram of Our Proposed CodeCytos Agent.} \textbf{a,} CodeCytos agent diagram, given the requested spatial feature by bioscientists, the agent carries out a multi-turn of thinking/reasoning, text/code generation, environment observation steps. This is based on the idea of ReAct agent \cite{yao2022react}, which iteratively thinks, acts, and observes. Moreover, CodeCytos action space is extended with the ability of writing code and execute code explicitly \cite{wang2024executable}, enabling the agent to leverage the programming knowledge acquired by LLMs during pretraining. \textbf{b,} The \textbf{\textcolor{Emerald}{Environment}} supplies contextual information that the CodeCytos agent uses for observation and subsequent reasoning. This context includes CodeCytos system prompt, the biologists’ requested spatial features, the selected cell-segmentation and cell-classification tools, the uploaded spatial cellular images, and the agent’s in-context working state (e.g., prior reasoning steps, generated code, execution logs, and any raised errors).
}
    
    \label{fig:CodeCytos_algorithm}
\end{figure}

In this work, we develop CodeCytos, a coding- and reasoning-based agent designed to support spatial molecular imaging analysis. We are motivated by a key limitation of conventional analysis softwares: while they offers built-in, pre-implemented functionality, they often provides limited support for defining and analyzing custom spatial cellular features. Our system follows the Reasoning–Acting (ReAct) agent design pattern \cite{yao2022react}, in which the agent’s action space is augmented with explicit “thought” steps represented as natural-language text generated by a large language model (LLM). However, because natural language constitutes an extremely large and unconstrained action space, selecting effective actions can be difficult. As a result, ReAct-style agents typically rely on substantial prompt engineering and expert-crafted few-shot demonstrations to adapt an LLM to a specific domain or task. To reduce this dependence on manual prompting, we expand CodeCytos’s action space with executable code, building on the CodeAct paradigm \cite{wang2024executable} (figure \ref{fig:CodeCytos_algorithm}). CodeAct unifies an agent’s actions within a single, code-based action space; by leveraging LLMs’ pretrained programming knowledge, it can perform effectively even in zero-shot settings without human demonstrations, thereby minimizing human effort \cite{wang2024executable}.

To evaluate CodeCytos, we assembled a benchmark dataset of tissue field-of-view images spanning multiple tissue types: frontal cortex, non-small cell lung cancer (NSCLC), pancreas, and tonsil. For each tissue type, we constructed a dedicated set of questions targeting cellular morphology and spatial organization, curated in consultation with an expert biologist. The benchmark is intentionally designed for a minimal prompt setting, in which questions are presented without task instructions or contextual information about the underlying spatial analysis. Preliminary results are promising: CodeCytos consistently outperforms baseline approaches, underscoring the practical potential of code-action agents for spatial cellular analysis. Notably, incorporating few-shot, guided coding-and-reasoning exemplars further improves performance, even when the exemplars are domain-agnostic (i.e., none are related to spatial molecular imaging analysis). This few-shot setting yields a significant boost over the original CodeAct in the zero-shot setting. Finally, we conducted extensive evaluations across multiple state-of-the-art coding-focused LLMs and observed consistent improvements when few-shot examples were provided. We further evaluated multiple state-of-the-art coding-oriented LLMs and observed consistent performance improvements with few-shot conditioning. We report Success Rate and Pass@k (k = 5, 10, 20), and introduce Area Under the Pass@k Curve (AUP@k), a summary metric that aggregates performance across Pass@k values for different k.

To better understand why out-of-domain, task-agnostic coding-and-reasoning exemplars can substantially improve CodeCytos on spatial molecular imaging tasks, we conducted the same experiment on $M^3$ToolEval \cite{wang2024executable}, the benchmark introduced in the original CodeAct paper for multi-turn interactions requiring multiple tool calls. Using the same few-shot exemplars as in our spatial cellular analysis benchmark, we observe that task-agnostic few-shot prompting also improves performance on $M^3$ToolEval. Although the gains are smaller than those observed on our spatial analysis benchmark, the improvement remains noticeable and outperforms training-free baselines, including text-as-action, JSON-as-action, and the original zero-shot CodeAct. These results further support the benefit of incorporating tailored few-shot coding-and-reasoning exemplars to improve CodeAct-style agent performance, particularly when using open-source LLM backbones.

\subsection*{ReAct, CodeAct, and Few-Shot In-Context Demonstrations}

\subsubsection*{ReAct and CodeAct Agent}
Consider an agent interacting with an environment to solve a task. At time step $t$, the agent receives an observation $o_t \in O$ and selects an action $a_t \in A$ according to a policy $\pi(a_t \mid c_t)$, where the context $c_t$ is
\begin{equation}
    c_t = (o_1, a_1, \ldots, o_{t-1}, a_{t-1}, o_t).    
\end{equation}

ReAct augments the agent's action space with language, yielding $\tilde{A} = A \cup L$, where $L$ denotes ``thought'' in text format generated by a Large Language Model (LLM) \cite{yao2022react}. Thought actions do not affect the environment and therefore produce no new observation; instead, they serve to organize information and perform reasoning over the current context $c_t$, updating the internal context for subsequent steps, e.g.,
\[
c_{t+1} = (c_t, \tilde{a}_t).
\]

However, language is vast, and thus language space is unlimited, making choosing agent actions based on context from this augmented action space very challenging. In ReAct \cite{yao2022react}, the LLM is prompted with few-shot demonstrations to steer generation toward domain-appropriate environment actions accompanied by free-form ``thought'' for task solving. Each demonstration is an expert trajectory of interleaved thought, action, and environment observations, typically constructed from samples in the training data. For tasks in which reasoning is central, ReAct alternates between generating thoughts and actions, resulting in multi-step thought-action-observation trajectories \cite{yao2022react}.

Given the limitations of heavy prompt engineering and manually crafted few-shot demonstrations for adapting LLMs to particular domains or tasks, the CodeAct framework proposes using executable Python code as the agent interface, consolidating an LLM agent’s behaviors into a unified action space that can scale across tasks with minimal human effort. This approach leverages the prevalence of code in modern LLM pre-training: CodeAct can leverage existing software packages to expand its functional action space, importing appropriate Python libraries to solve tasks without requiring user-provided tools or in-context demonstrations. Compared with rigid JSON or fixed-format text actions, code natively supports control and data flow, enabling intermediate results to be stored as variables for reuse and allowing multiple operations (e.g., conditionals and loops) to be composed within a single program. This better unlocks LLMs’ ability to solve complex tasks by leveraging their pre-trained programming knowledge.

\subsubsection*{Few-shot Coding-Reasoning Prompting for In-Context Demonstrations}
In the original ReAct paper \cite{yao2022react}, actions are expressed in natural language, with tool calls hand-crafted into textual formats. The authors rely on few-shot in-context demonstrations by prepending the prompt with exemplars drawn from the training set. For knowledge-intensive reasoning, they sample six HotPotQA \cite{yang2018hotpotqa} instances and three FEVER \cite{thorne2018fever} instances, and manually write ReAct-style trajectories containing multiple interleaved thought–action–observation steps. Similarly, for ALFWorld \cite{shridhar2020alfworld} decision-making benchmark (covering six task types), they manually annotate three training trajectories per task type to serve as demonstrations.

Prior approaches often on extensive prompt engineering and carefully crafted few-shot demonstrations to adapt LLMs to specific domains or tasks \cite{yao2022react, liang2022code}, since base models are not explicitly optimized for dynamic planning and decision-making \cite{wang2024executable}. In ReAct, few-shot in-context examples are used to elicit both domain-specific actions and free-form “thoughts” that support task completion. However, producing agent rollout reasoning-trace annotations at scale is costly and challenging \cite{yao2022react}. An alternative is to train agents on gold ReAct-style trajectories—using successful task completion (e.g., matching ground-truth answers) as supervision—but this approach is computationally expensive and often limited by the availability of high-quality trajectory data.
Unlike ReAct, which encodes tool-use actions as handcrafted text in the language space—and therefore often requires substantial prompt engineering and few-shot demonstrations to adapt an LLM to a specific domain or task—CodeAct \cite{wang2024executable} proposes executable Python code as the agent interface, unifying an agent’s actions within a single, code-based action space. By leveraging LLMs’ strong prior knowledge of programming, CodeAct-style agents can generalize across diverse tasks and domains with minimal human effort. Empirically, CodeAct reports notable improvements over prior baselines even in the zero-shot setting. However, the authors also report a substantial zero-shot performance gap between open-source and proprietary models: the best open-source model achieves 13.4\%, whereas the best closed-source model reaches 74.4\%.

Although CodeAct reports that open-source models lag behind proprietary models in the zero-shot setting, our experiments show that several open-source backbones can be substantially improved by prepending a small number of coding-and-reasoning exemplars for few-shot in-context learning. Importantly, these exemplars are domain-agnostic—general programming demonstrations unrelated to spatial cellular analysis, yet they still yield significant performance gains.

\subsubsection*{Effectiveness of Few-shot In-context learning in improving agent}

While reinforcement learning–based methods have become a common go-to for improving agent behavior \cite{nemo-gym, cheng2025agentr1trainingpowerfulllm, deepswe2025, rafailov2023direct, guo2025deepseek}, in-context learning (ICL) remains a low-cost, practical alternative without the need for training—especially when weight updates are infeasible due to implementation or deployment constraints. As a model-agnostic approach, ICL enables the same prompting strategy to transfer across LLM backbones and is well-suited to low-data settings where collecting agent trajectory traces is expensive. Moreover, theoretical and empirical evidence suggests that ICL performance scales with additional demonstrations \cite{akyurek2022learning, von2023transformers, bertsch2025context, agarwal2024many}, implying that adding a small set of few-shot exemplars can yield substantial gains.

\section*{Results}

\subsection*{Dataset Information}

\subsubsection*{Cellular Spatial Tissue Dataset}
We assembled a diverse set of fields of view (FOVs) across multiple tissue types to ensure broad coverage of cellular compositions and microenvironments. All tissue FOVs were obtained from a public repository \cite{nanostring_website}. The dataset comprises FOVs from four tissues: Frontal Cortex, NSCLC, Pancreas, and Tonsil. For each tissue, domain experts formulated targeted questions spanning a wide range of cellular characteristics, including neighbor-based features, cell morphology, and spatial organization. We intentionally evaluate in a minimal-prompt regime: questions are presented without task instructions or additional contextual hints about the underlying spatial analysis, reducing the influence of prompt engineering and emphasizing whether a method can infer the appropriate analysis from the question and execute it reliably. This question set is intended to capture tissue-specific biological insights reflecting distinct cell types and microenvironmental context. Tables \ref{tab:spatial_features_enumerate} and \ref{tab:feature_distribution} summarize the distribution and frequency of feature categories across tissues and provide definitions, highlighting the complexity and heterogeneity inherent in spatial cellular data analysis.

\begingroup
\setlength{\LTleft}{0pt plus 1fill}
\setlength{\LTright}{0pt plus 1fill}
\scriptsize 

\begin{longtable}{|P{0.18\linewidth}|P{0.30\linewidth}|P{0.22\linewidth}|P{0.25\linewidth}|}
\caption{\textbf{Categorization of spatial cellular features in the expert-curated dataset.} The table summarizes an abstract taxonomy of spatial descriptors that quantify tissue organization from cell coordinates. Features are grouped into five classes: \textbf{(1) NND \& distances}, nearest-neighbor/\(k\)-NN proximity to cells or landmarks and summaries of distance distributions, including neighbor-type preferences; \textbf{(2) local neighborhoods}, fixed-radius microenvironment measures (counts/ratios and rule-based proximity) together with global FOV composition and diversity/entropy summaries; \textbf{(3) spatial statistics}, point-process functions and indices that characterize clustering or dispersion relative to a null hypothesis model across spatial scales; \textbf{(4) structural \& ROI}, descriptors tied to larger tissue structures, including cluster/partition geometry and positional, shape, and orientation features; and \textbf{(5) graph \& topology}, graph-derived measures summarizing edge geometry, connectivity, and, when available, interface/contact properties.}
\label{tab:spatial_features_enumerate} \\
\hline
\textbf{Primary Category} & \textbf{Description} & \textbf{Feature Subgroups} & \textbf{Examples} \\
\hline
\endfirsthead

\multicolumn{4}{c}%
{{\tablename\ \thetable{} -- continued from previous page}} \\
\hline
\textbf{Primary Category} & \textbf{Description} & \textbf{Feature Subgroups} & \textbf{Examples} \\
\hline
\endhead

\hline
\multicolumn{4}{|r|}{{Continued on next page}} \\
\hline
\endfoot

\hline
\endlastfoot

\textbf{1. NND \& Distances} & Metrics based on the Euclidean distance from a cell's centroid to its closest neighbor(s) or a distant landmark. This includes measurements of shortest distance ($k$-NN) and statistical analysis of the resulting distance distributions (CV, preference). & \begin{itemize}[nosep, itemsep=0pt, leftmargin=*]
    \item \textbf{Basic Proximity:} Measures distance to the very first nearest neighbor (NN);
    \item \textbf{k-NN Distances:} Measures distance to the $k^{th}$ nearest neighbor ($k > 1$)
    \item \textbf{Derived NND Stats:} Statistical properties (spread, skew) of the NND distribution
    \item \textbf{NN Type Preference:} Identifies the cell type of the nearest neighbor. 
\end{itemize}    
    &
\begin{enumerate}[label=\roman*., nosep, itemsep=0pt, leftmargin=*]
    \item Mean NND from T cells to the closest epithelial cell.
    \item Proportion of T cells whose nearest neighbor is Epithelial.
    \item $90^{th}$ percentile of NNDs; $\text{CV}$ of NNDs.
    \item Proportion of T cells whose nearest neighbor is Epithelial.
\end{enumerate} \\
\hline

\textbf{2. Local Neighborhoods} & Focuses on the composition or count of cells within a fixed-radius circular neighborhood, simulating local microenvironments. & \textbf{A. Local Composition and Context:} Quantifies the counts, ratios, or complex rule fulfillment within a local radius. 
\begin{itemize}[nosep, itemsep=0pt, leftmargin=*]
    \item \textbf{Proportions/Fractions:} Binary check: the percentage of cells that have at least one neighbor within a radius.
    \item \textbf{Counts \& Ratios:} Mean or median count of one cell type around another; local ratios.
    \item \textbf{Complex Rules/Triads:} Questions involving nested rules, exclusion, or diversity indices.
\end{itemize}
 &
\begin{enumerate}[label=\roman*., nosep, itemsep=0pt, leftmargin=*]
    \item Proportion of epithelial cells with at least one T cell within $30\,\mu\text{m}$.
    \item Mean count of stromal cells within $20\,\mu\text{m}$.
    \item Odds ratio for a T cell to have an epithelial neighbor.
\end{enumerate} \\
\cline{3-4}
& & \textbf{B. Global Composition and Diversity:} Simple summary statistics for the entire Field of View (FOV).

\begin{itemize}[nosep, itemsep=0pt, leftmargin=*]
    \item \textbf{Global/Basic Composition:} Simple counts or ratios for the entire FOV (density/composition).
    \item \textbf{Diversity/Entropy:} quantifying the mix, richness, or disorder of cell types, either within localized neighborhoods or globally across the Field of View (FOV).
\end{itemize}
&
\begin{enumerate}[label=\roman*., nosep, itemsep=0pt, leftmargin=*]
    \item Global T-cell density ($\text{cells}/\text{mm}^2$).
    \item Astrocyte fraction among all cells.
    \item Median Shannon Diversity Index of Cell Types within $50\,\mu\text{m}$ neighborhoods centered on Macrophages.
\end{enumerate} \\
\hline

\textbf{3. Spatial Stats} & Uses statistical functions (Spatial Point Process Theory) to test a specific spatial hypothesis (e.g., clustering, dispersion) by comparing observed cell patterns against a null model (e.g., $\text{CSR}$). This includes pairwise functions ($\text{Ripley's K}$) and grid/global indices ($\text{Moran's I}$). & 
\begin{itemize}[nosep, itemsep=0pt, leftmargin=*]
    \item \textbf{Pairwise Functions:} Measures how cell interactions change over a range of distances ($r$).
    \item \textbf{Global Indices:} Single-value summary statistics for the overall cells organization pattern of clustering or dispersal.
    \item \textbf{Distance Distribution Functions:} Measures based on the distribution of distances to the first nearest event or empty space.
    \item \textbf{Quadrat/Gridding:} Metrics derived from dividing the FOV into bins and analyzing cell counts per bin.
\end{itemize}
&
\begin{enumerate}[label=\roman*., nosep, itemsep=0pt, leftmargin=*]
    \item $\text{Cross-K}\ L_{12}(r)$ function for T cells and Epithelial cells.
    \item Pair correlation $g(r)$.
    \item Clark–Evans $R$ for T cells.
    \item $\text{Moran's}\ I$ for T cell intensity.
    \item $G\text{-function}\ G(r)$; $F\text{-function}\ F(r)$; $J\text{-function}\ J(r)$.
    \item Variance-to-Mean Ratio (VMR).
    \item Maximum cell count in any $100\,\mu\text{m}$ tile.
    
\end{enumerate} \\
\hline

\textbf{4. Structural \& ROI} & Focuses on geometry, shape, and a cell's position relative to large, pre-defined tissue landmarks or clusters. & \textbf{A. Structure and Segmentation:} Metrics describing the size, number, or area of structures identified by clustering algorithms ($\text{DBSCAN}$) or space-dividing methods ($\text{Voronoi}$ Tessellation).

\begin{itemize}[nosep, itemsep=0pt, leftmargin=*]
    \item \textbf{Cluster Properties:} Metrics describing the size, number, or composition of groups identified by algorithms (e.g., DBSCAN).
    \item \textbf{Tessellation \& Area:} Features derived from Voronoi tiling or cell segmentation area/overlap.
\end{itemize}

&
\begin{enumerate}[label=\roman*., nosep, itemsep=0pt, leftmargin=*]
    \item Mean size of T-cell clusters defined by $\text{DBSCAN}$.
    \item Median T-cell $\text{Voronoi cell area}$.
    \item $\text{Convex hull area}$ of the largest cluster.
\end{enumerate} \\
\cline{3-4}
& & \textbf{B. Positional, Shape, and Directional Features:} Metrics related to cell shape/orientation ($\text{PCA}, \text{Fractal Dimension}$) and a cell's position relative to FOV boundaries or specific cluster boundaries ($\text{Signed Distance}$). 

\begin{itemize}[nosep, itemsep=0pt, leftmargin=*]
    \item \textbf{Boundary \& Edge Effects:} Cell position relative to the FOV edge or cluster boundaries.
    \item \textbf{Morphology \& Orientation:} Metrics related to the shape or orientation of cell segmentations.
\end{itemize}

&
\begin{enumerate}[label=\roman*., nosep, itemsep=0pt, leftmargin=*]
    \item $\text{Fractal dimension}$ of T-cell segmentation.
    \item $\text{Signed distance}$ to epithelial cluster boundary.
    \item Median angle of major axis relative to neighbor.
\end{enumerate} \\
\hline

\textbf{5. Graph \& Topology} & Constructs a network where cells are nodes and connections are edges (based on rules like Delaunay or k-NN), then quantifies properties related to the network foundation, overall connectivity, and contact/interface properties. & 

\begin{itemize}[nosep, itemsep=0pt, leftmargin=*]
    \item \textbf{Graph Construction:} Metrics related to the foundational edges of graphs (Delaunay, MST, Gabriel).
    \item \textbf{Network Connectivity:} Measures of how well-connected or dense the entire network is.
    \item \textbf{Contact/Interface Metrics:} Metrics focused on physical polygon-to-polygon contact.
\end{itemize}

&
\begin{enumerate}[label=\roman*., nosep, itemsep=0pt, leftmargin=*]
    \item $\text{Mean edge length}$ of $\text{Delaunay}$ triangulation.
    \item Fraction of mixed-type edges in kNN graph.
    \item $\text{Clustering Coefficient}$ of the k-NN graph.
    \item Size of the Largest Connected Component.
    \item Mean $\text{length of shared boundary}$ between cells.
\end{enumerate} \\
\hline
\end{longtable}
\endgroup

\begingroup
\setlength{\LTleft}{0pt plus 1fill}
\setlength{\LTright}{0pt plus 1fill}
\scriptsize 

\begin{longtable}{|P{0.22\linewidth}|P{0.28\linewidth}|c|c|c|c|}
\caption{\textbf{Distribution of spatial cellular feature categories across datasets.} Counts indicate the number of expert-curated features assigned to each primary spatial category (and, where applicable, subgroup) in four datasets (NSCLC, Frontal\_Cortex, Pancreas, Tonsil). Categories include nearest-neighbor distance–based measures (NND/\(k\)-NN and derived summaries), fixed-radius local neighborhoods and global FOV composition/diversity, point-process spatial statistics, structural/ROI geometry and positional/morphological descriptors, and graph/topology features. The final row reports per-dataset totals across all categories.}
\label{tab:feature_distribution} \\
\hline
\textbf{Primary Category} & \textbf{Feature Subgroups} & \textbf{NSCLC} & \textbf{Frontal\_Cortex} & \textbf{Pancreas} & \textbf{Tonsil} \\
\hline
\endfirsthead

\multicolumn{6}{c}%
{{\tablename\ \thetable{} -- continued from previous page}} \\
\hline
\textbf{Primary Category} & \textbf{Subtype (if applicable)} & \textbf{NSCLC} & \textbf{Frontal\_Cortex} & \textbf{Pancreas} & \textbf{Tonsil} \\
\hline
\endhead

\hline
\multicolumn{6}{|r|}{{Continued on next page}} \\
\hline
\endfoot

\hline
\endlastfoot

\textbf{1. NND \& Distances} & 
\begin{itemize}[nosep, itemsep=0pt, leftmargin=*]
    \item \textbf{Basic Proximity}
    \item \textbf{k-NN Distances}
    \item \textbf{Derived NND Stats}
    \item \textbf{NN Type Preference}
\end{itemize} 
& 34 & 10 & 22 & 19 \\
\hline

\textbf{2. Local Neighborhoods} & \textbf{A. Local Composition and Context} 
\begin{itemize}[nosep, itemsep=0pt, leftmargin=*]
    \item \textbf{Proportions/Fractions}
    \item \textbf{Counts \& Ratios}
    \item \textbf{Complex Rules/Triads}
\end{itemize}
& 52 & 30 & 59 & 48 \\
\cline{2-6}
& \textbf{B. Global Composition and Diversity} 

\begin{itemize}[nosep, itemsep=0pt, leftmargin=*]
    \item \textbf{Global/Basic Composition}
    \item \textbf{Diversity/Entropy}
\end{itemize}

& 2 & 6 & 20 & 12 \\
\hline

\textbf{3. Spatial Stats} & 

\begin{itemize}[nosep, itemsep=0pt, leftmargin=*]
    \item \textbf{Pairwise Functions}
    \item \textbf{Global Indices}
    \item \textbf{Distance Distribution Functions}
    \item \textbf{Quadrat/Gridding}
\end{itemize}

& 20 & 41 & 26 & 27 \\
\hline

\textbf{4. Structural \& ROI} & \textbf{A. Structure and Segmentation} 

\begin{itemize}[nosep, itemsep=0pt, leftmargin=*]
    \item \textbf{Cluster Properties}
    \item \textbf{Tessellation \& Area}
\end{itemize}

& 11 & 9 & 2 & 9 \\
\cline{2-6}
& \textbf{B. Positional, Shape, and Directional Features} 

\begin{itemize}[nosep, itemsep=0pt, leftmargin=*]
    \item \textbf{Boundary \& Edge Effects}
    \item \textbf{Morphology \& Orientation}
\end{itemize}

& 9 & 16 & 9 & 6 \\
\hline

\textbf{5. Graph \& Topology} & 

\begin{itemize}[nosep, itemsep=0pt, leftmargin=*]
    \item \textbf{Graph Construction}
    \item \textbf{Network Connectivity}
    \item \textbf{Contact/Interface Metrics}
\end{itemize}

& 6 & 12 & 21 & 10 \\
\hline

\hline
\textbf{Total} & & \textbf{134} & \textbf{124} & \textbf{159} & \textbf{131} \\
\hline

\end{longtable}
\endgroup

\subsubsection*{$M^3$ToolEval Dataset}

To verify whether adding task-agnostic few-shot guided programming exemplars will help on the original CodeAct benchmark, we also carry out experiments on $M^3$ToolEval \cite{wang2024executable}, the benchmarking dataset curated in the original CodeAct paper and the foundation on which our CodeCytos agent builds. $M^3$ToolEval evaluates an LLM’s ability to solve complex, multi-step tasks that require invoking multiple tools over multi-turn interactions. It contains 82 manually curated instances spanning domains such as web browsing, finance, travel planning, science, and information processing, with each domain is provided with a custom toolset \cite{wang2024executable}.

\subsection*{Agent Setup Overview}

Our agent is implemented using the CodeAct framework \cite{wang2024executable}, which extends the ReAct \cite{yao2022react} paradigm by allowing the model to generate and execute code as actions within a multi-step reasoning process. In this paper, for all ReAct-style experiments, each agent run is capped at 10 steps; the agent terminates either when it produces a final answer or when the step limit is reached. In this work, we define a run as a single interaction that begins with a bioscientist’s question and ends when the agent outputs a final answer or reaches the step limit. A run is considered successful if the final answer matches the ground truth; otherwise, it is counted as a failure.

\subsection*{Agent Validation and Evaluation Metrics}

To evaluate the agent, we report two primary metrics: success rate and pass@k \cite{chen2021evaluating}. Success rate is the fraction of questions answered correctly in a single run. Unless otherwise specified, we use the terms run and attempt interchangeably. Pass@k measures whether the agent produces at least one correct solution within k independent attempts, capturing robustness under stochastic sampling and the likelihood of success given multiple tries. We report pass@5, pass@10, and pass@20. We also report the area under the pass@k curve (AUP@k) as a summary statistic that aggregates pass@k across different values of k into a single, interpretable value. Together, these metrics provide a robust evaluation of the effectiveness and reliability of our code-as-action agent on our expert-curated spatial cellular analysis benchmark.

\subsection*{Agent Evaluation on Cellular Spatial Tissue dataset}

\subsubsection*{Overall Evaluation Results}
The two tables \ref{tab:CodeCytos-frontal-nsclc-results} and \ref{tab:CodeCytos-pancreas-tonsil-results} shows evaluation results for the code–action agent across different datasets. Each experiment differs in the choice of LLM backbone; we select models with strong coding ability, including Qwen2.5-Coder-32B-Instruct \cite{hui2024qwen2}, Qwen3-Coder-30B-A3B-Instruct \cite{yang2025qwen3}, Facebook-Code World Model (CWM) \cite{copet2025cwm}, Devstral-Small-2505 \cite{rastogi2025devstral}, Devstral-2-123-Instruct-2512 \cite{rastogi2025devstral}, Kimi-Linear-48B-A3B-Instruct \cite{team2025kimi}, and GLM-4.5-Air \cite{zeng2025glm}.

For each dataset, we evaluate two primary settings: (1) a tool-augmented LLM with access to a Python interpreter, the uploaded image, and precomputed cell segmentation and classification outputs; and (2) our proposed CodeAct-based \textsc{CodeCytos} agent. Within the tool-augmented LLM setting, we consider three variants: (a) a conventional tool-augmented baseline, (b) a zero-shot chain-of-thought (CoT) \cite{wei2022chain} variant, and (c) a few-shot CoT variant. Within the \textsc{CodeCytos} setting, we likewise evaluate three variants: (a) a conventional CodeAct-based \textsc{CodeCytos} baseline, (b) a one-shot CodeAct variant, and (c) a few-shot CodeAct variant.

Our best-performing configuration, the few-shot CodeAct-based \textsc{CodeCytos}, uses several domain-agnostic coding-and-reasoning exemplars to steer the agent’s reasoning and code generation. Across all datasets, this setting consistently outperforms other variants, achieving higher success rates, pass@k, and AUP@k. These results underscore the value of few-shot guided coding–reasoning exemplars for in-context learning: despite being task-agnostic, they help the CodeAct agent execute the step-by-step coding–reasoning cycle more effectively.

\begin{figure}[htbp]
  \centering

  \begin{subfigure}[b]{0.48\textwidth}
    \centering
    \includegraphics[width=\linewidth]{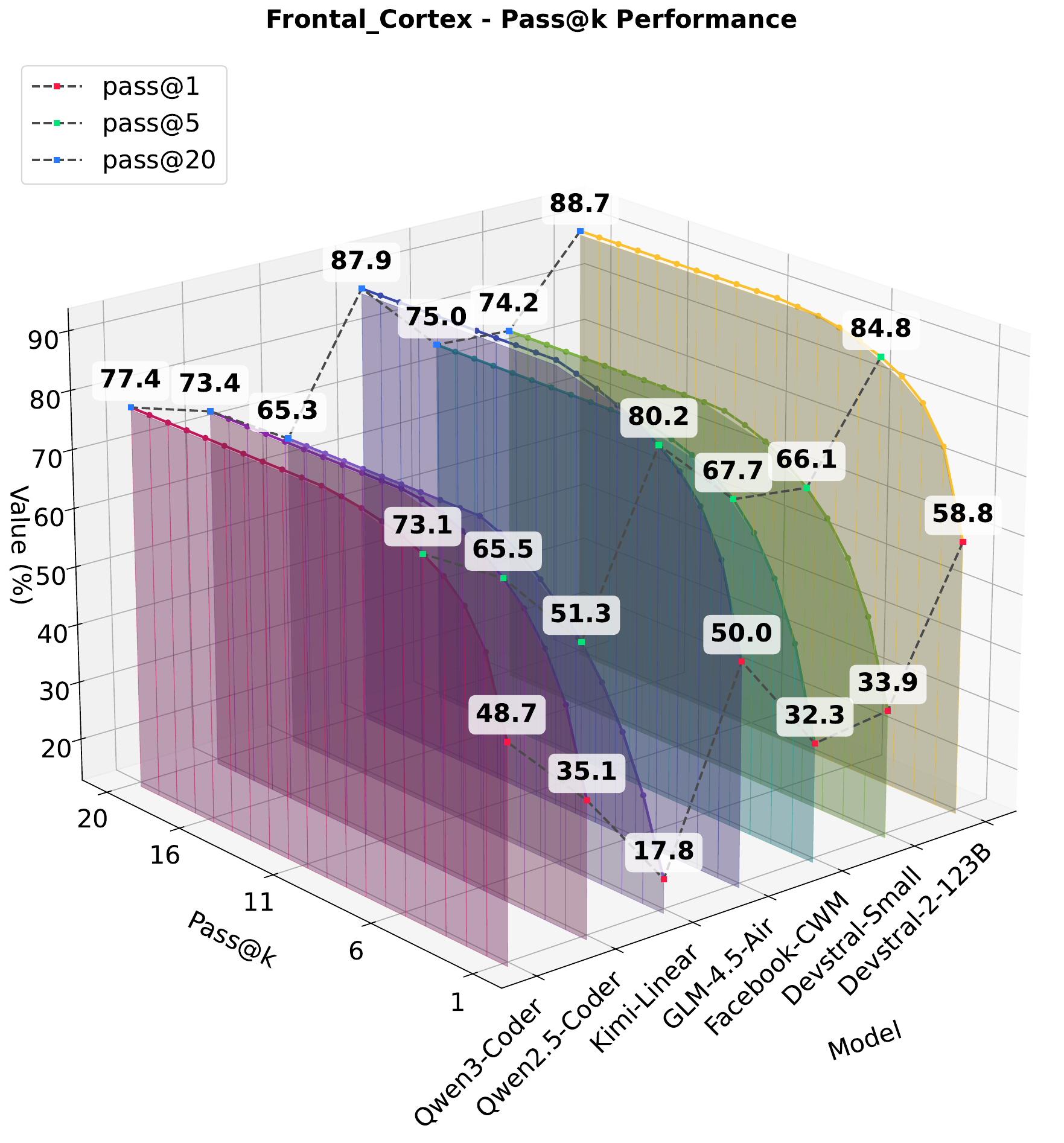}
    \caption{Frontal Cortex}
    \label{fig:sub1}
  \end{subfigure}\hfill
  \begin{subfigure}[b]{0.48\textwidth}
    \centering
    \includegraphics[width=\linewidth]{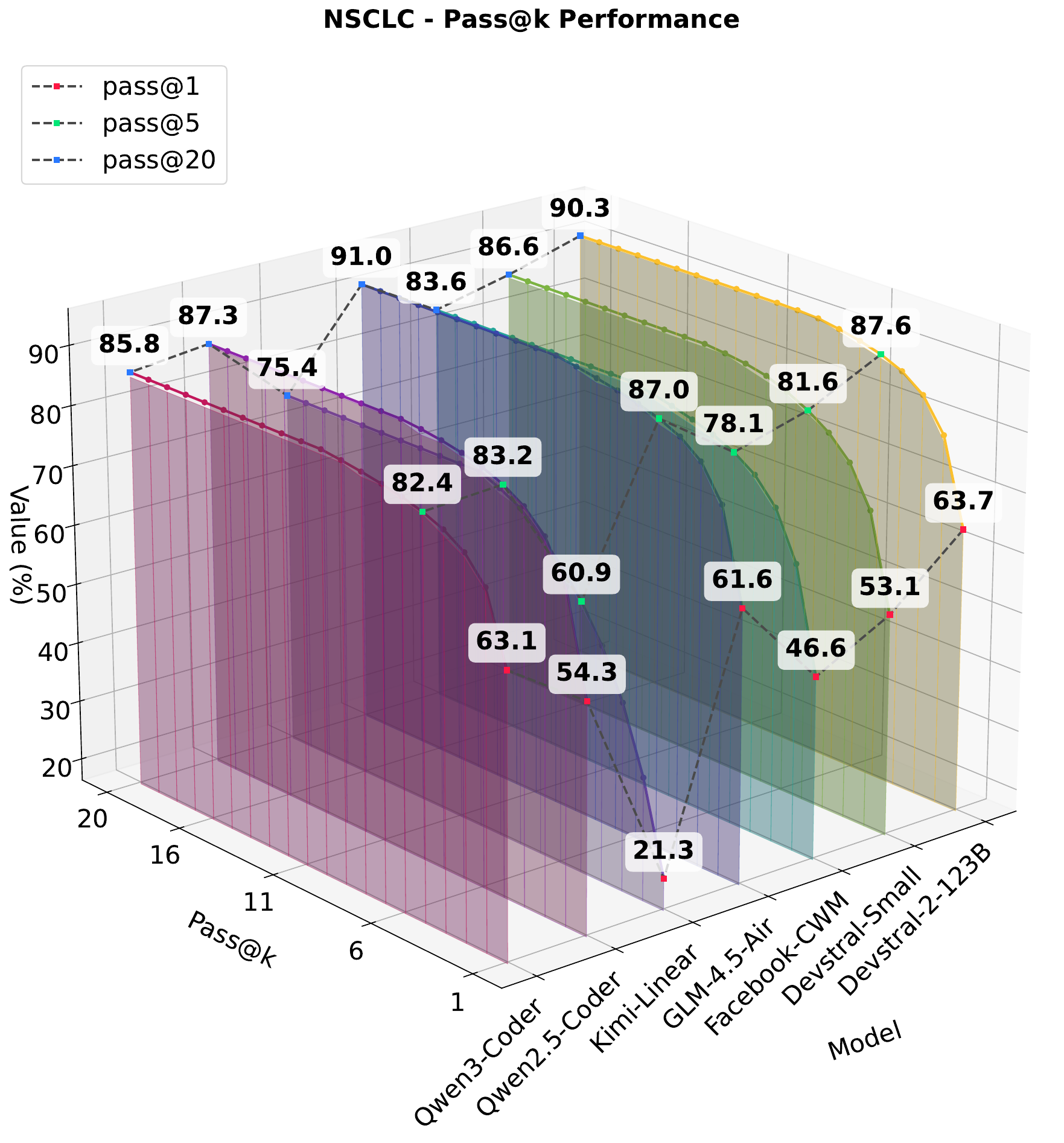}
    \caption{Non-Small Cell Lung Cancer}
    \label{fig:sub2}
  \end{subfigure}

  \vspace{0.5em}

  \begin{subfigure}[b]{0.48\textwidth}
    \centering
    \includegraphics[width=\linewidth]{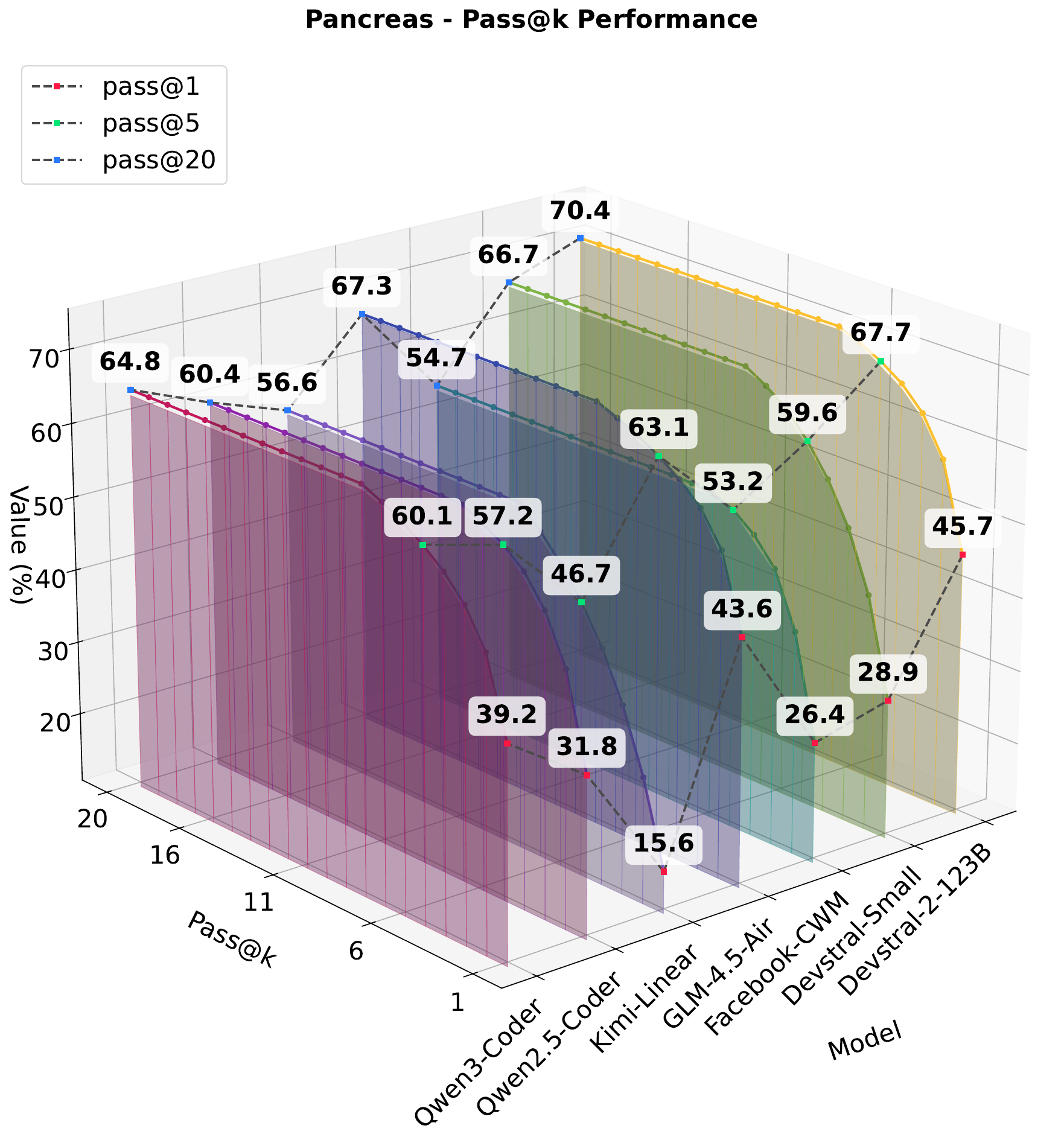}
    \caption{Pancreas}
    \label{fig:sub3}
  \end{subfigure}\hfill
  \begin{subfigure}[b]{0.48\textwidth}
    \centering
    \includegraphics[width=\linewidth]{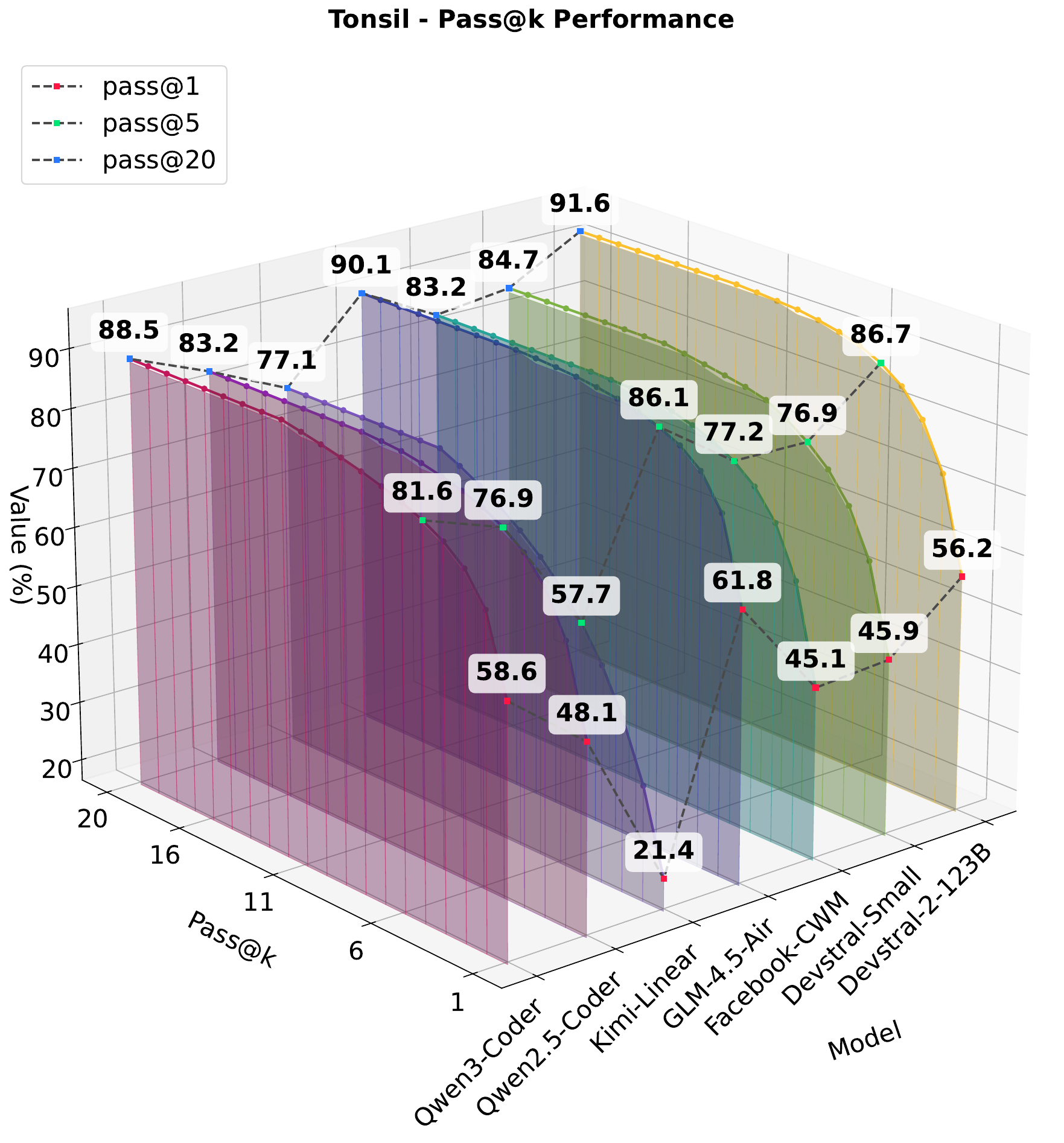}
    \caption{Tonsil}
    \label{fig:sub4}
  \end{subfigure}

  \caption{\textbf{Pass@k curves for CodeCytos across four tissue-type datasets with different LLM backbones.} The choice of LLM backbone leads to distinct pass@k trajectories across datasets. Performance generally begins to plateau around $k\approx10$; we therefore report representative values at $k\in\{1,5,20\}$. Overall, GLM-4.5-Air and Devstral-2-123B achieve the strongest results. While increasing the number of attempts raises pass@k, the curves indicate that additional attempts alone do not guarantee perfect success, motivating future work to further improve CodeCytos reliability.}
  \label{fig:pass@k_curves_all_datasets}
\end{figure}

\textbf{Pass@k curves analysis:} We further analyze CodeCytos by plotting pass@k curves for different LLM backbones across four tissue-type datasets using our best setting-CodeCytos with few-shot exemplars (Fig.~\ref{fig:pass@k_curves_all_datasets}). The curves differ substantially across backbones, indicating that the choice of underlying model strongly affects how effectively additional attempts translate into successful solutions. Overall, pass@k increases with $k$, with the largest gains at small $k$ (e.g., $k=1$ to $k=5$) and diminishing returns thereafter. In line with this trend, performance typically begins to plateau around $k \approx 10$; we therefore report representative values at $k \in \{1,5,20\}$ to capture both early and near-saturated performance.

Overall, GLM-4.5-Air and Devstral-2-123B achieve the strongest results across datasets, showing higher pass@k throughout the range of $k$ and converging to higher plateaus than the other backbones. Importantly, while increasing the number of attempts generally improves success rates, the plateauing behavior suggests that additional attempts alone do not guarantee perfect performance: some failures persist even at larger $k$. This motivates future work on improving CodeCytos reliability more systematically, for example via stronger planning and tool-use policies, improved verification or self-correction, and training strategies that target recurring failure modes rather than relying on repeated sampling.

\begingroup
\setlength{\LTleft}{0pt plus 1fill}
\setlength{\LTright}{0pt plus 1fill}
\setlength{\tabcolsep}{0.2pt}
\scriptsize

\begin{longtable}{|l|cc|cc|cc|cc|cc|cc|}

\caption{\textbf{CodeCytos vs. baselines on Frontal\_Cortex and NSCLC.} CodeCytos augmented with task-agnostic few-shot coding--reasoning demonstrations achieves the strongest performance across settings. For each LLM backbone, we report five metrics: success\_rate, pass@5, pass@10, pass@20, and AUP@k. For NSCLC, \redbar[45] and \redbar[15] denote the best and second-best value per metric, respectively; for Frontal\_Cortex, \bluebar[45] and \bluebar[15] denote the best and second-best value per metric.}
\label{tab:CodeCytos-frontal-nsclc-results} \\
\hline

\textbf{Metric}
& \multicolumn{6}{c|}{\textbf{Tool-Augmented LLM}}
& \multicolumn{6}{c|}{\textbf{CodeAct-based CodeCytos}} \\
\hline

& \multicolumn{2}{c|}{\textbf{\makecell{Conventional\\ Without CoT}}}
& \multicolumn{2}{c|}{\textbf{\makecell{With\\ Zero-Shot CoT}}}
& \multicolumn{2}{c|}{\textbf{\makecell{With\\ Few-Shot CoT}}}
& \multicolumn{2}{c|}{\textbf{CodeCytos}}
& \multicolumn{2}{c|}{\textbf{\makecell{CodeCytos\\+ One Shot}}}
& \multicolumn{2}{c|}{\textbf{\makecell{CodeCytos\\+ Few Shot}}} \\
\hline

\textbf{Dataset}
 & \rotatebox{90}{\textcolor{blue!60!black}{Frontal\_Cortex}} & \rotatebox{90}{\textcolor{red!70!black}{NSCLC}}
 & \rotatebox{90}{\textcolor{blue!60!black}{Frontal\_Cortex}} & \rotatebox{90}{\textcolor{red!70!black}{NSCLC}}
 & \rotatebox{90}{\textcolor{blue!60!black}{Frontal\_Cortex}} & \rotatebox{90}{\textcolor{red!70!black}{NSCLC}}
 & \rotatebox{90}{\textcolor{blue!60!black}{Frontal\_Cortex}} & \rotatebox{90}{\textcolor{red!70!black}{NSCLC}}
 & \rotatebox{90}{\textcolor{blue!60!black}{Frontal\_Cortex}} & \rotatebox{90}{\textcolor{red!70!black}{NSCLC}}
 & \rotatebox{90}{\textcolor{blue!60!black}{Frontal\_Cortex}} & \rotatebox{90}{\textcolor{red!70!black}{NSCLC}} \\
\hline
\endfirsthead

\multicolumn{13}{c}{{\tablename\ \thetable{} -- continued from previous page}} \\
\hline

\textbf{Metric}
& \multicolumn{6}{c|}{\textbf{Tool-Augmented LLM}}
& \multicolumn{6}{c|}{\textbf{CodeAct-based CodeCytos}} \\
\hline

& \multicolumn{2}{c|}{\textbf{\makecell{Conventional\\ Without CoT}}}
& \multicolumn{2}{c|}{\textbf{\makecell{With\\ Zero-Shot CoT}}}
& \multicolumn{2}{c|}{\textbf{\makecell{With\\ Few-Shot CoT}}}
& \multicolumn{2}{c|}{\textbf{CodeCytos}}
& \multicolumn{2}{c|}{\textbf{\makecell{CodeCytos\\+ One Shot}}}
& \multicolumn{2}{c|}{\textbf{\makecell{CodeCytos\\+ Few Shot}}} \\
\hline

\textbf{Dataset}
 & \rotatebox{90}{\textcolor{blue!60!black}{Frontal\_Cortex}} & \rotatebox{90}{\textcolor{red!70!black}{NSCLC}}
 & \rotatebox{90}{\textcolor{blue!60!black}{Frontal\_Cortex}} & \rotatebox{90}{\textcolor{red!70!black}{NSCLC}}
 & \rotatebox{90}{\textcolor{blue!60!black}{Frontal\_Cortex}} & \rotatebox{90}{\textcolor{red!70!black}{NSCLC}}
 & \rotatebox{90}{\textcolor{blue!60!black}{Frontal\_Cortex}} & \rotatebox{90}{\textcolor{red!70!black}{NSCLC}}
 & \rotatebox{90}{\textcolor{blue!60!black}{Frontal\_Cortex}} & \rotatebox{90}{\textcolor{red!70!black}{NSCLC}}
 & \rotatebox{90}{\textcolor{blue!60!black}{Frontal\_Cortex}} & \rotatebox{90}{\textcolor{red!70!black}{NSCLC}} \\
\hline
\endhead

\hline
\multicolumn{13}{|r|}{{Continued on next page}} \\
\hline
\endfoot

\hline
\endlastfoot

\multicolumn{13}{|l|}{\textbf{Qwen2.5-Coder-32B-Instruct}} \\
\multicolumn{13}{|l|}{\textit{\quad Year: 2024 \hspace{2cm} Size: 32.5 Bilions Parameters \hspace{2cm} Pretrain Data: 5.5 trillion of training tokens }} \\
\hline
success\_rate & 2.35 & \cellcolor{red!15}6.07 & 2.60 & 5.69 & \cellcolor{blue!15}3.95 & 5.97 & 0.75 & 0.60 & 0.16 & 0.45 & \cellcolor{blue!45}36.61 & \cellcolor{red!45}54.28 \\
{\tiny CI} & {\tiny [1.40, 3.39]} & \cellcolor{red!15}{\tiny [4.03, 8.43]} & {\tiny [1.51, 3.87]} & {\tiny [3.81, 7.84]} & \cellcolor{blue!15}{\tiny [2.45, 5.65]} & {\tiny [4.03, 8.21]} & {\tiny [0.33, 1.24]} & {\tiny [0.22, 1.12]} & {\tiny [0.00, 0.40]} & {\tiny [0.15, 0.82]} & \cellcolor{blue!45}{\tiny [30.87, 42.33]} & \cellcolor{red!45}{\tiny [49.10, 59.55]} \\
\cline{1-13}
pass@5 & 10.52 & 19.62 & 10.53 & 19.30 & \cellcolor{blue!15}14.66 & \cellcolor{red!15}20.34 & 3.72 & 2.80 & 0.80 & 2.26 & \cellcolor{blue!45}65.56 & \cellcolor{red!45}83.24 \\
{\tiny CI} & {\tiny [6.72, 14.70]} & {\tiny [14.40, 25.19]} & {\tiny [6.43, 15.18]} & {\tiny [14.29, 24.77]} & \cellcolor{blue!15}{\tiny [9.76, 19.87]} & \cellcolor{red!15}{\tiny [15.06, 25.92]} & {\tiny [1.61, 6.14]} & {\tiny [1.12, 5.06]} & {\tiny [0.00, 2.02]} & {\tiny [0.75, 4.10]} & \cellcolor{blue!45}{\tiny [58.12, 72.77]} & \cellcolor{red!45}{\tiny [77.45, 88.69]} \\
\cline{1-13}
pass@10 & 17.63 & 29.85 & 16.13 & 29.93 & \cellcolor{blue!15}21.82 & \cellcolor{red!15}30.58 & 7.27 & 5.20 & 1.59 & 4.51 & \cellcolor{blue!45}73.45 & \cellcolor{red!45}87.26 \\
{\tiny CI} & {\tiny [11.29, 24.21]} & {\tiny [22.39, 38.06]} & {\tiny [9.68, 22.58]} & {\tiny [22.39, 38.06]} & \cellcolor{blue!15}{\tiny [14.52, 29.03]} & \cellcolor{red!15}{\tiny [23.13, 38.81]} & {\tiny [3.23, 12.10]} & {\tiny [1.49, 8.96]} & {\tiny [0.00, 4.03]} & {\tiny [1.49, 8.21]} & \cellcolor{blue!45}{\tiny [65.32, 80.65]} & \cellcolor{red!45}{\tiny [81.34, 92.54]} \\
\hline
\end{longtable}
\endgroup

\begingroup
\setlength{\LTleft}{0pt plus 1fill}
\setlength{\LTright}{0pt plus 1fill}
\setlength{\tabcolsep}{0.2pt}
\scriptsize

\begin{longtable}{|l|cc|cc|cc|cc|cc|cc|}
\caption{\textbf{CodeCytos vs. baselines on Pancreas and Tonsil.} CodeCytos augmented with task-agnostic few-shot coding--reasoning demonstrations achieves the strongest performance across settings. For each LLM backbone, we report five metrics: success\_rate, pass@5, pass@10, pass@20, and AUP@k. For Pancreas, \violetbar[45] and \violetbar[15] denote the best and second-best value per metric, respectively; For Tonsil, \orangebar[45] and \orangebar[15] denote the best and second-best value per metric.}
\label{tab:CodeCytos-pancreas-tonsil-results} \\
\hline

\textbf{Metric}
& \multicolumn{6}{c|}{\textbf{Tool-Augmented LLM}}
& \multicolumn{6}{c|}{\textbf{CodeAct-based CodeCytos}} \\
\hline

& \multicolumn{2}{c|}{\textbf{\makecell{Conventional\\ Without CoT}}}
& \multicolumn{2}{c|}{\textbf{\makecell{With\\ Zero-Shot CoT}}}
& \multicolumn{2}{c|}{\textbf{\makecell{With\\ Few-Shot CoT}}}
& \multicolumn{2}{c|}{\textbf{CodeCytos}}
& \multicolumn{2}{c|}{\textbf{\makecell{CodeCytos\\+ One Shot}}}
& \multicolumn{2}{c|}{\textbf{\makecell{CodeCytos\\+ Few Shot}}} \\
\hline

\textbf{Dataset}
 & \rotatebox{90}{\textcolor{violet!60!black}{Pancreas}} & \rotatebox{90}{\textcolor{orange!70!black}{Tonsil}}
 & \rotatebox{90}{\textcolor{violet!60!black}{Pancreas}} & \rotatebox{90}{\textcolor{orange!70!black}{Tonsil}}
 & \rotatebox{90}{\textcolor{violet!60!black}{Pancreas}} & \rotatebox{90}{\textcolor{orange!70!black}{Tonsil}}
 & \rotatebox{90}{\textcolor{violet!60!black}{Pancreas}} & \rotatebox{90}{\textcolor{orange!70!black}{Tonsil}}
 & \rotatebox{90}{\textcolor{violet!60!black}{Pancreas}} & \rotatebox{90}{\textcolor{orange!70!black}{Tonsil}}
 & \rotatebox{90}{\textcolor{violet!60!black}{Pancreas}} & \rotatebox{90}{\textcolor{orange!70!black}{Tonsil}} \\
\hline
\endfirsthead

\multicolumn{13}{c}{{\tablename\ \thetable{} -- continued from previous page}} \\
\hline

\textbf{Metric}
& \multicolumn{6}{c|}{\textbf{Tool-Augmented LLM}}
& \multicolumn{6}{c|}{\textbf{CodeAct-based CodeCytos}} \\
\hline

& \multicolumn{2}{c|}{\textbf{\makecell{Conventional\\ Without CoT}}}
& \multicolumn{2}{c|}{\textbf{\makecell{With\\ Zero-Shot CoT}}}
& \multicolumn{2}{c|}{\textbf{\makecell{With\\ Few-Shot CoT}}}
& \multicolumn{2}{c|}{\textbf{CodeCytos}}
& \multicolumn{2}{c|}{\textbf{\makecell{CodeCytos\\+ One Shot}}}
& \multicolumn{2}{c|}{\textbf{\makecell{CodeCytos\\+ Few Shot}}} \\
\hline

\textbf{Dataset}
 & \rotatebox{90}{\textcolor{violet!60!black}{Pancreas}} & \rotatebox{90}{\textcolor{orange!70!black}{Tonsil}}
 & \rotatebox{90}{\textcolor{violet!60!black}{Pancreas}} & \rotatebox{90}{\textcolor{orange!70!black}{Tonsil}}
 & \rotatebox{90}{\textcolor{violet!60!black}{Pancreas}} & \rotatebox{90}{\textcolor{orange!70!black}{Tonsil}}
 & \rotatebox{90}{\textcolor{violet!60!black}{Pancreas}} & \rotatebox{90}{\textcolor{orange!70!black}{Tonsil}}
 & \rotatebox{90}{\textcolor{violet!60!black}{Pancreas}} & \rotatebox{90}{\textcolor{orange!70!black}{Tonsil}}
 & \rotatebox{90}{\textcolor{violet!60!black}{Pancreas}} & \rotatebox{90}{\textcolor{orange!70!black}{Tonsil}} \\
\hline
\endhead

\hline
\multicolumn{13}{|r|}{{Continued on next page}} \\
\hline
\endfoot

\hline
\endlastfoot

\multicolumn{13}{|l|}{\textbf{Qwen2.5-Coder-32B-Instruct}} \\
\multicolumn{13}{|l|}{\textit{\quad Year: 2024 \hspace{2cm} Size: 32.5 Bilions Parameters \hspace{2cm} Pretrain Data: 5.5 trillion of training tokens }} \\
\hline
success\_rate & \cellcolor{violet!15}4.02 & 4.65 & 4.01 & 5.62 & 3.78 & \cellcolor{orange!15}7.17 & 0.62 & 0.89 & 0.63 & 1.53 & \cellcolor{violet!45}32.44 & \cellcolor{orange!45}47.65 \\
{\tiny CI} & \cellcolor{violet!15}{\tiny [2.67, 5.50]} & {\tiny [3.24, 6.17]} & {\tiny [2.59, 5.58]} & {\tiny [3.88, 7.51]} & {\tiny [2.52, 5.19]} & \cellcolor{orange!15}{\tiny [5.09, 9.48]} & {\tiny [0.24, 1.10]} & {\tiny [0.51, 1.34]} & {\tiny [0.16, 1.18]} & {\tiny [0.95, 2.16]} & \cellcolor{violet!45}{\tiny [27.21, 37.51]} & \cellcolor{orange!45}{\tiny [41.97, 53.49]} \\
\cline{1-13}
pass@5 & \cellcolor{violet!15}15.14 & 18.33 & 14.18 & 20.51 & 14.85 & \cellcolor{orange!15}23.22 & 3.16 & 4.45 & 2.69 & 7.40 & \cellcolor{violet!45}57.22 & \cellcolor{orange!45}76.88 \\
{\tiny CI} & \cellcolor{violet!15}{\tiny [10.61, 20.17]} & {\tiny [13.63, 23.37]} & {\tiny [9.57, 19.14]} & {\tiny [15.49, 25.72]} & {\tiny [10.39, 19.73]} & \cellcolor{orange!15}{\tiny [17.63, 29.14]} & {\tiny [1.18, 5.50]} & {\tiny [2.23, 6.68]} & {\tiny [0.79, 5.00]} & {\tiny [4.66, 10.38]} & \cellcolor{violet!45}{\tiny [49.96, 64.68]} & \cellcolor{orange!45}{\tiny [70.30, 82.97]} \\
\cline{1-13}
pass@10 & 20.15 & 29.04 & 18.26 & 32.12 & \cellcolor{violet!15}20.76 & \cellcolor{orange!15}33.13 & 5.04 & 8.91 & 3.75 & 14.22 & \cellcolor{violet!45}60.32 & \cellcolor{orange!45}82.52 \\
{\tiny CI} & {\tiny [13.84, 27.04]} & {\tiny [21.93, 36.48]} & {\tiny [12.58, 24.53]} & {\tiny [24.95, 39.81]} & \cellcolor{violet!15}{\tiny [14.47, 27.04]} & \cellcolor{orange!15}{\tiny [25.76, 40.86]} & {\tiny [1.89, 8.81]} & {\tiny [4.45, 13.36]} & {\tiny [1.26, 6.92]} & {\tiny [9.02, 19.55]} & \cellcolor{violet!45}{\tiny [52.20, 67.92]} & \cellcolor{orange!45}{\tiny [75.90, 88.76]} \\
\hline
\end{longtable}
\endgroup

\subsubsection*{Evaluation Results on Separate Spatial Features Categories}

\begin{figure}[htbp]
    \centering
    \includegraphics[width=\textwidth]{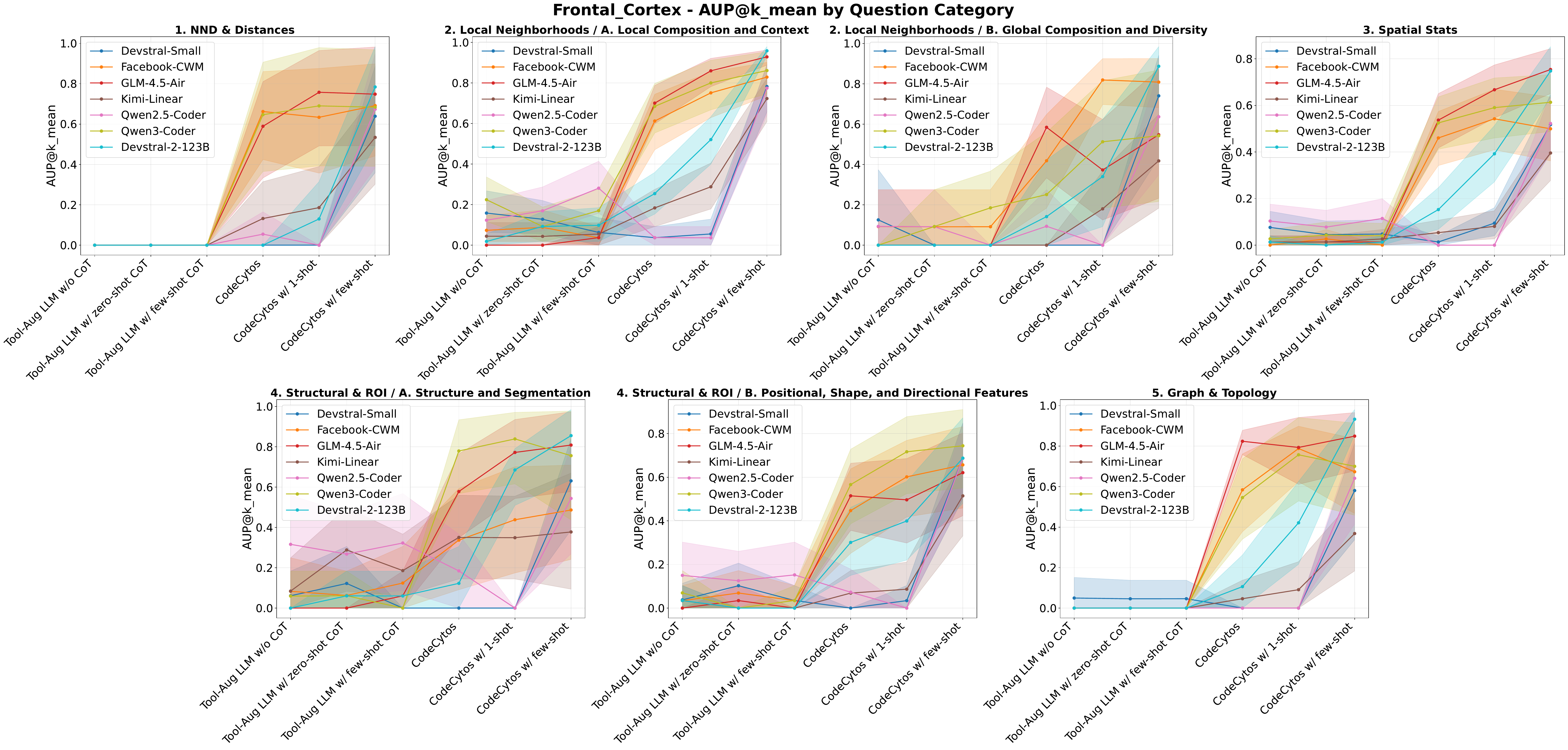}
    \caption{\textbf{AUP@k by category on the Frontal\_Cortex dataset.} We compare two settings: (1) a tool-augmented LLM and (2) the CodeAct-based CodeCytos agent. The first three x-axis groups correspond to tool-augmented LLM variants: (1.a) without CoT, (1.b) zero-shot CoT, and (1.c) few-shot CoT. The last three x-axis groups correspond to CodeCytos variants: (2.a) CodeCytos, (2.b) CodeCytos with 1-shot demonstrations, and (2.c) CodeCytos with few-shot demonstrations. Overall, CodeCytos outperforms the conventional tool-augmented LLM, and adding domain-agnostic few-shot coding--reasoning exemplars further improves performance.}
    \label{fig:frontal_cortex_aupk}
\end{figure}

\begin{figure}[htbp]
    \centering
    \includegraphics[width=\textwidth]{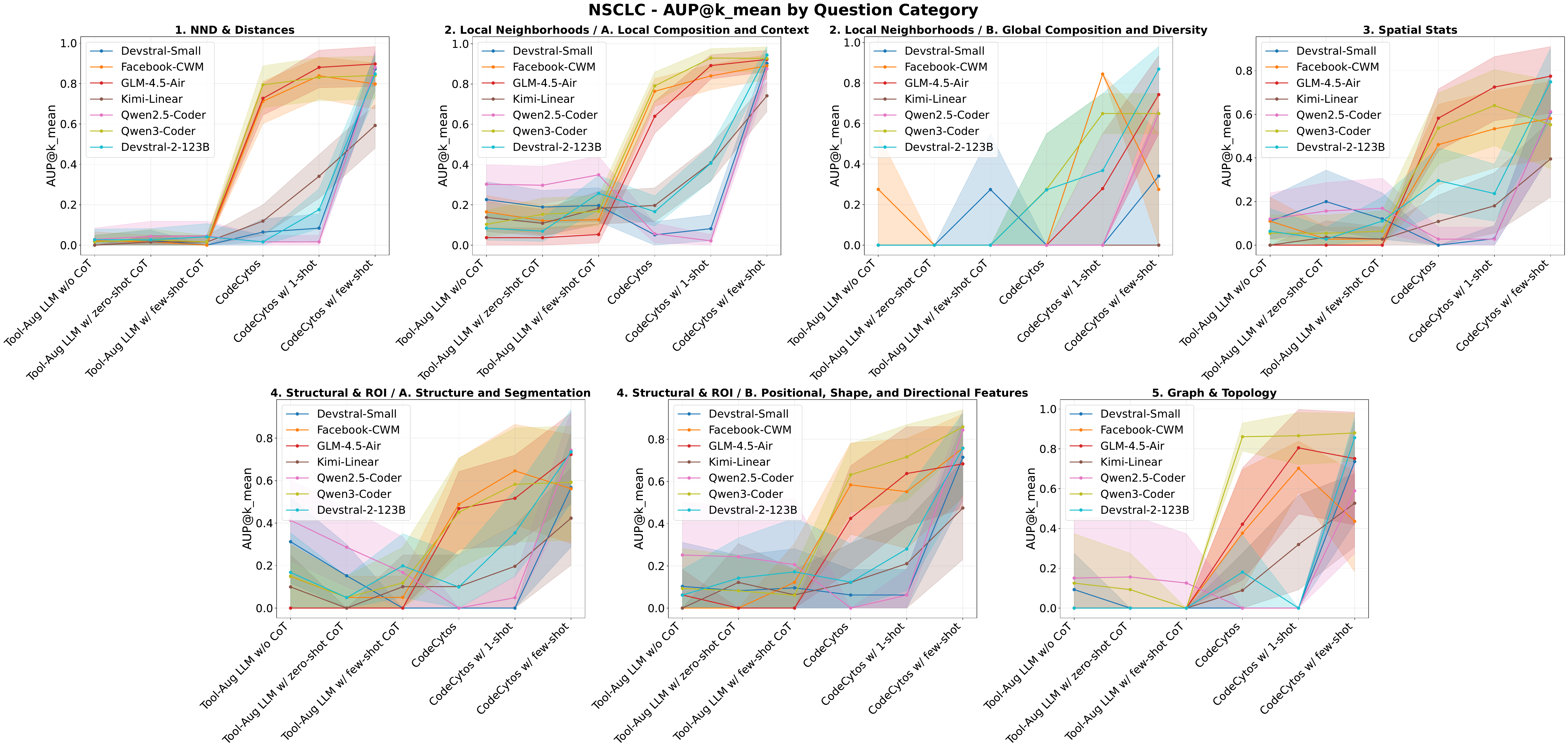}
   \caption{\textbf{AUP@k by category on the NSCLC dataset.} We compare two settings: (1) a tool-augmented LLM and (2) the CodeAct-based CodeCytos agent. The first three x-axis groups correspond to tool-augmented LLM variants: (1.a) without CoT, (1.b) zero-shot CoT, and (1.c) few-shot CoT. The last three x-axis groups correspond to CodeCytos variants: (2.a) CodeCytos, (2.b) CodeCytos with 1-shot demonstrations, and (2.c) CodeCytos with few-shot demonstrations. Overall, CodeCytos outperforms the conventional tool-augmented LLM, and adding domain-agnostic few-shot coding--reasoning exemplars further improves performance.}
    \label{fig:nsclc_aupk}
\end{figure}

\begin{figure}[htbp]
    \centering
    \includegraphics[width=\textwidth]{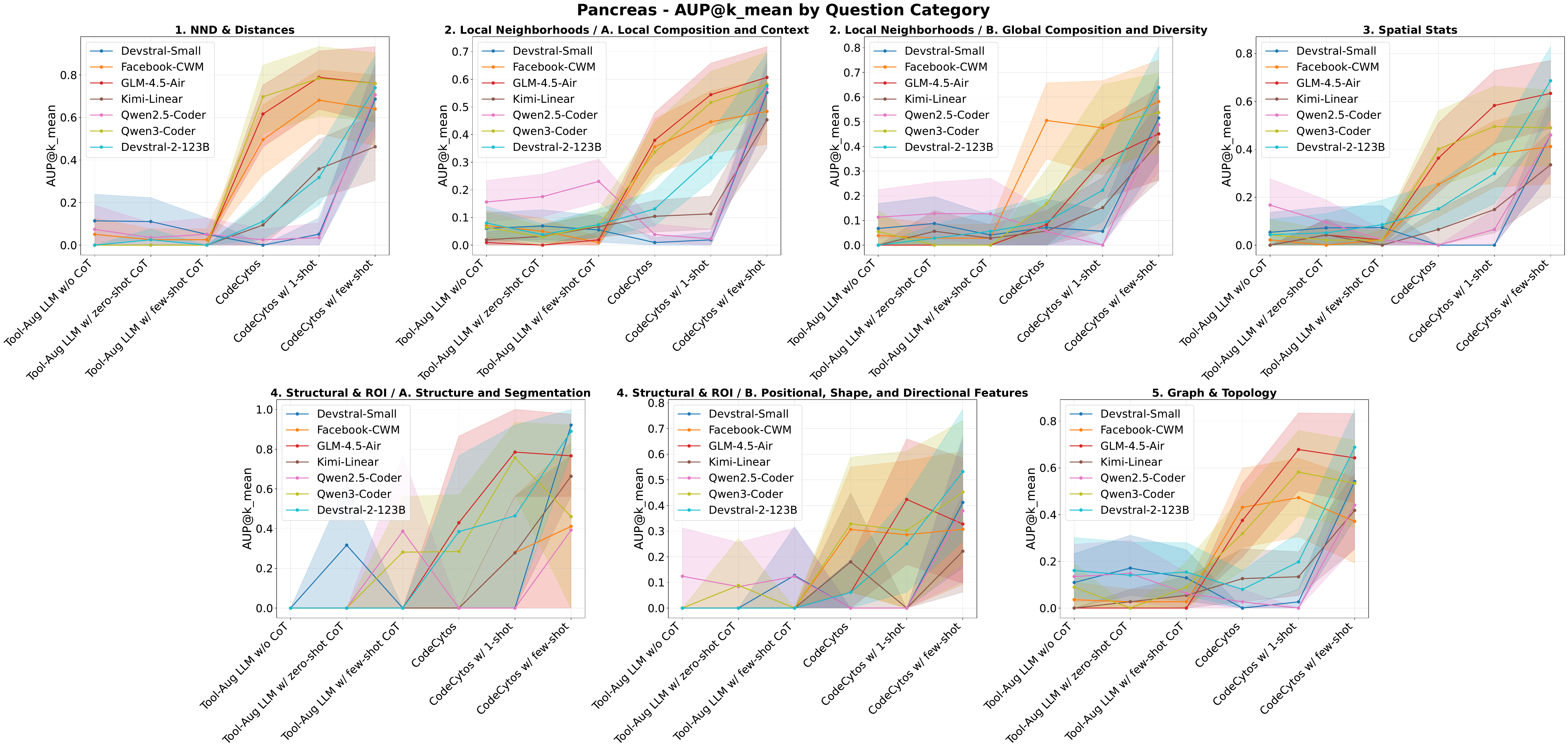}
   \caption{\textbf{AUP@k by category on the Pancreas dataset.} We compare two settings: (1) a tool-augmented LLM and (2) the CodeAct-based CodeCytos agent. The first three x-axis groups correspond to tool-augmented LLM variants: (1.a) without CoT, (1.b) zero-shot CoT, and (1.c) few-shot CoT. The last three x-axis groups correspond to CodeCytos variants: (2.a) CodeCytos, (2.b) CodeCytos with 1-shot demonstrations, and (2.c) CodeCytos with few-shot demonstrations. Overall, CodeCytos outperforms the conventional tool-augmented LLM, and adding domain-agnostic few-shot coding--reasoning exemplars further improves performance.}
    \label{fig:pancreas_aupk}
\end{figure}

\begin{figure}[htbp]
    \centering
    \includegraphics[width=\textwidth]{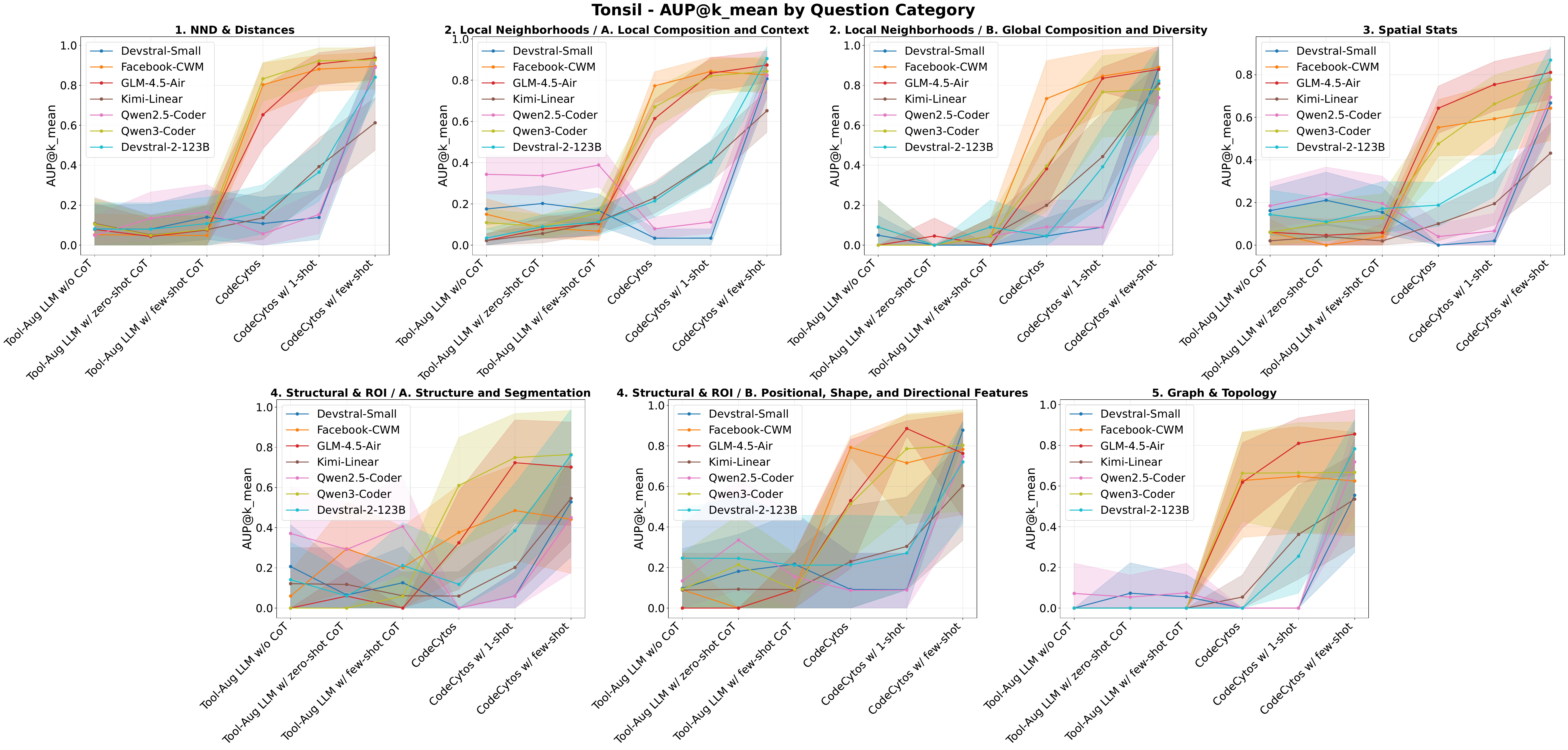}
   \caption{\textbf{AUP@k by category on the Tonsil dataset.} We compare two settings: (1) a tool-augmented LLM and (2) the CodeAct-based CodeCytos agent. The first three x-axis groups correspond to tool-augmented LLM variants: (1.a) without CoT, (1.b) zero-shot CoT, and (1.c) few-shot CoT. The last three x-axis groups correspond to CodeCytos variants: (2.a) CodeCytos, (2.b) CodeCytos with 1-shot demonstrations, and (2.c) CodeCytos with few-shot demonstrations. Overall, CodeCytos outperforms the conventional tool-augmented LLM, and adding domain-agnostic few-shot coding--reasoning exemplars further improves performance.}
    \label{fig:tonsil_aupk}
\end{figure}

We further evaluate performance across spatial feature categories—nearest-neighbor distance, local neighborhood, spatial statistics, structural \& ROIs, and graph-based features—to characterize category-specific variations. Results for each feature category and dataset are shown in figures \ref{fig:frontal_cortex_aupk}, \ref{fig:nsclc_aupk}, \ref{fig:pancreas_aupk}, and \ref{fig:tonsil_aupk} for Frontal Cortex, NSCLC, Pancreas, and Tonsil, respectively. Overall, the few-shot CodeAct setting with guided coding–reasoning exemplars achieves the highest performance across feature categories for all datasets, followed by the one-shot CodeAct setting.

We also observe trend differences across coding-capable LLM backbones. For comparison, we consider two main settings, each further divided into three variants: (a) a tool-augmented LLM setting, including a tool-augmented baseline, a zero-shot CoT variant, and a few-shot CoT variant; and (b) a CodeAct-based \textsc{CodeCytos} setting, including a conventional CodeAct-based \textsc{CodeCytos} baseline, a one-shot exemplar variant, and a few-shot exemplar variant. Across models, we observe distinct performance patterns.

Compared with the other backbones, Qwen3-A30-Coder-Instruct, Facebook-CWM, and GLM-4.5-Air consistently perform better under the iterative CodeAct-based CodeCytos setting across zero-shot, one-shot, and few-shot variants. Even in the zero-shot variant, these models achieve strong performance; adding guided coding-reasoning exemplars generally further improves results across spatial feature categories. There are a few exceptions where guided exemplars do not help. For example, on the \textit{Frontal Cortex} dataset for the \textit{Global Compositions and Diversity} task, GLM-4.5-Air does not improve with exemplars and shows a slight decrease. Similarly, on the \textit{Pancreas} and \textit{Tonsil} datasets for the \textit{Positional, Shape, and Directional Features} task and the \textit{Graph and Topology} task, Facebook-CWM shows little to no improvement, and in some cases a slight decrease, when provided with guided exemplars.

Devstral-2-123B, Devstral-small-2505, Kimi-Linear-48B-A3B-Instruct, and Qwen2.5-Coder-32B-Instruct all derive substantial gains from guided coding–reasoning exemplars. Devstral-small-2505 and Qwen2.5-Coder-32B-Instruct, despite typically performing well in tool-augmented LLM settings, degrade sharply under Zero-Shot and One-Shot CodeAct and recover only when provided with few-shot coding–reasoning demonstrations, highlighting the role of guided exemplars in eliciting their CodeAct capabilities. Kimi-Linear-48B-A3B-Instruct exhibits a different pattern, achieving higher accuracy in the Zero-Shot CodeAct setting and improving further with a one-shot exemplar, yet underperforms Devstral-small-2505 and Qwen2.5-Coder-32B-Instruct in the few-shot setting. Finally, Devstral-2-123B—the strongest model in our experiments—consistently benefits from coding–reasoning exemplars: while one-shot CodeAct improves over its own zero-shot setting, Devstral-2-123B in zero-shot and one-shot CodeAct still lags significantly behind those of Facebook-CWM, Qwen3-Coder, and GLM-4.5-Air. Only with the addition of few-shot exemplars does performance increase substantially, bringing it in line with, or even surpassing the previously leading models. This further underscores the importance of incorporating few-shot coding–reasoning demonstrations.

Overall, incorporating few-shot coding–reasoning exemplars improves performance across models, datasets, and tasks. Although the magnitude of the gains varies with the LLM backbone and task, the results highlight the importance of using tailored exemplars to guide code-action agents. Notably, equipping our CodeAct-based CodeCytos agent with a small set of domain-agnostic few-shot coding–reasoning exemplars (i.e., not specific to spatial analysis) substantially boosts performance, making this few-shot setting consistently the best across all evaluated configurations.

\subsubsection*{Performance Analyses across Different Models and Features Categories}

\begin{figure}[htbp]
    \centering
    
    \includegraphics[width=\textwidth]{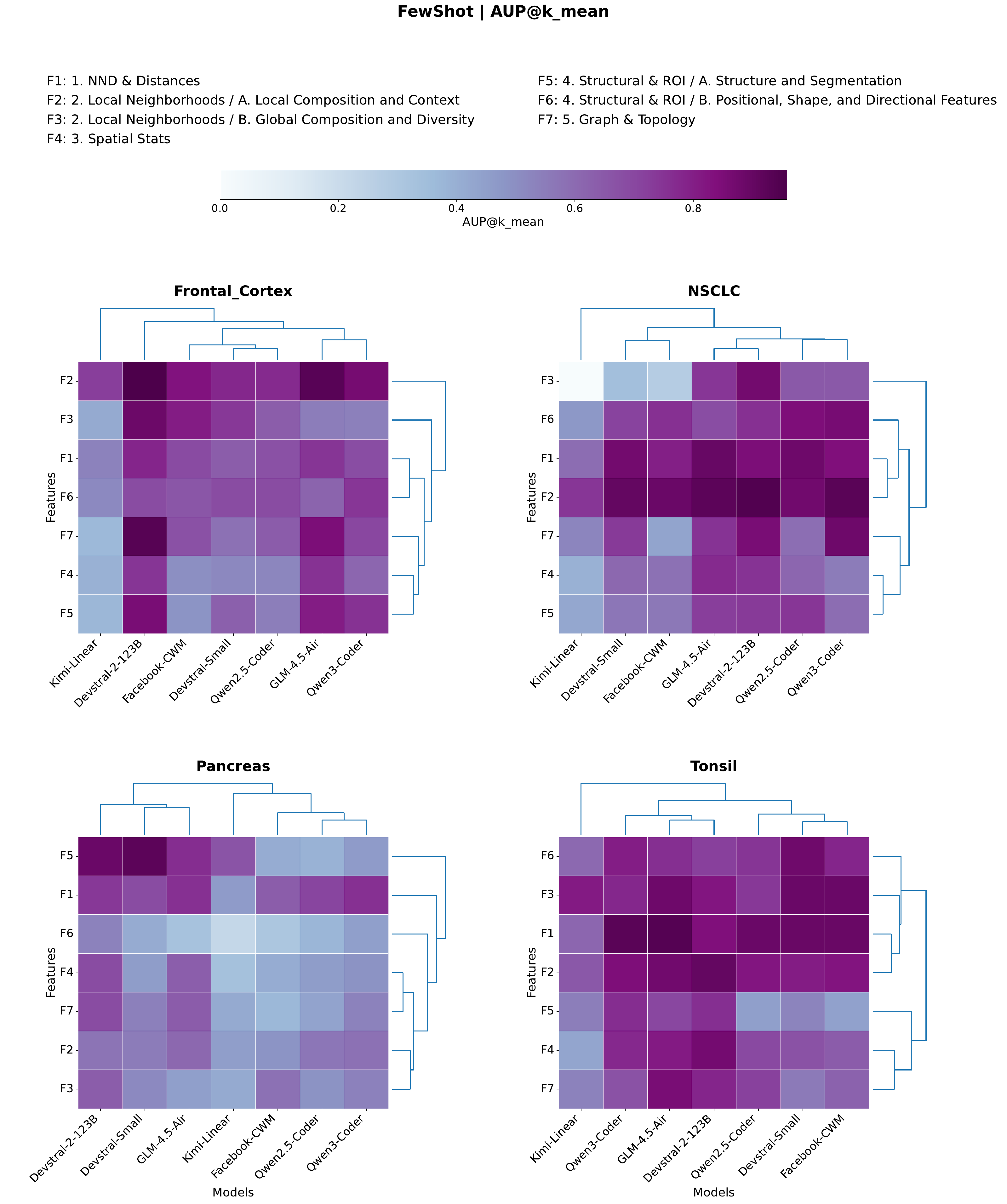}    

    \caption{\textbf{Performance heatmaps across four datasets.} Heatmaps compare multiple LLM backbones across feature categories and datasets, revealing that Nearest Neighbor Distances (NND) consistently achieve the highest performance and appear least challenging, with neighborhood-related features typically following. Models cluster into two groups—Kimi-Linear-Instruct vs. coding-optimized backbones—with Devstral-2-123B achieving the highest overall performance and GLM-4.5-Air ranking second. Task clustering patterns are broadly consistent across datasets (notably NSCLC and Tonsil), where Structural/ROI, NND/distances, and local-neighborhood tasks form tight high-performing groups, underscoring dataset- and backbone-dependent variability.}
    \label{fig:features_categories_heatmap}
\end{figure}

Figure \ref{fig:features_categories_heatmap} presents performance heatmaps for multiple models and feature categories across four datasets. Across all datasets and LLM backbones, features based on Nearest Neighbor Distances (NND) consistently achieve the strongest results. Among the evaluated models, Devstral-2-123B attains the highest overall performance, as expected given that it is the largest model in our study and has been reported to be competitive on a range of coding benchmarks. GLM-4.5-Air ranks second, consistent with its scale and extensive pretraining on trillions of tokens.

Analysis of the heatmaps across the four datasets indicates that the models generally form two primary clusters. The first cluster contains only Kimi-Linear-Instruct, whereas the second includes Devstral-2-123B, GLM-4.5-Air, Qwen-2.5-Coder-32B-Instruct, and Qwen-3-Coder-32B-Instruct. This separation is expected: Kimi-Linear-Instruct is optimized for efficient long-context, general-purpose use via a hybrid attention mechanism rather than for coding, and therefore tends to underperform relative to models explicitly trained for coding tasks. Consistent with this pattern, the strongest coding model in our study—Devstral-2-123B—often appears as a leaf-level subcluster on its own or clusters with GLM-4.5-Air, reflecting its consistently higher performance; GLM-4.5-Air typically follows closely, as suggested by similarly intense heatmap coloration.

The heatmap analysis across the four datasets indicates that the Nearest Neighbor Distances (NND) category is the least challenging across all datasets, followed closely by neighborhood-feature tasks, which are the second least difficult. The remaining categories show greater variability, largely driven by the specific model and dataset.

In the Frontal Cortex dataset, the Graph Theory, Spatial Statistics, and Structural/Segmentation categories form a tight cluster, and Devstral-2-123B, GLM-4.5-Air, and Qwen3-Coder exhibit stronger performance than the other models. Similar patterns are observed in the NSCLC and Tonsil datasets. In addition, for NSCLC and Tonsil, the dendrogram reveals three tightly clustered task groups: (1) Structural and ROIs (positional, shape, and directional features), (2) NND and distances, and (3) local neighborhoods (local composition and context). These groupings consistently achieve the highest performance across models.

Overall, performance is strongly influenced by both dataset characteristics and task complexity. Moreover, using different coding LLM backbones introduces additional variability across tasks and datasets, suggesting opportunities for future work on methods that improve performance more consistently.

\subsubsection*{Human-Agent Interaction Examples for Cellular Spatial Tissue Analyses}

\begin{figure}
    \centering
    \includegraphics[width=\linewidth]{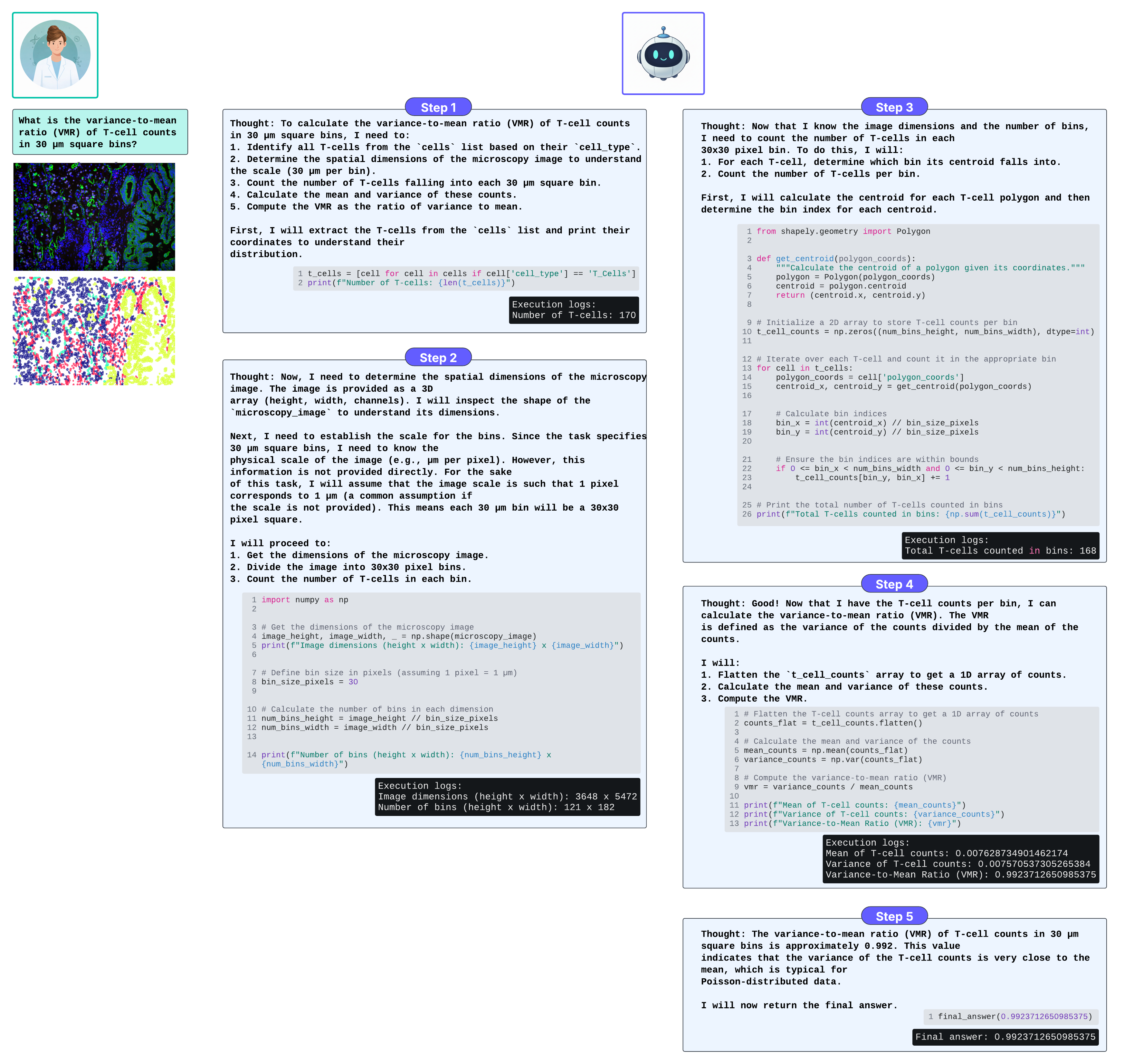}
    \caption{\textbf{Example CodeCytos agent run on an NSCLC tissue field of view (FOV) answering a bioscientist’s query: “What is the variance-to-mean ratio (VMR) of T-cell counts in 30~\(\mu\)m square bins?”} The bioscientist provides exactly the question shown in the figure, with no additional task specifications or spatial-analysis instructions. Notably, in step~2 the agent recognizes that the physical scale is not provided and therefore assumes a resolution of 1~\(\mu\)m. This suggests that the CodeCytos agent’s underlying LLM has acquired microscopy-imaging knowledge during pretraining.}
    \label{fig:chat_example_1}
\end{figure}

\begin{figure}
    \centering
    \includegraphics[width=\linewidth]{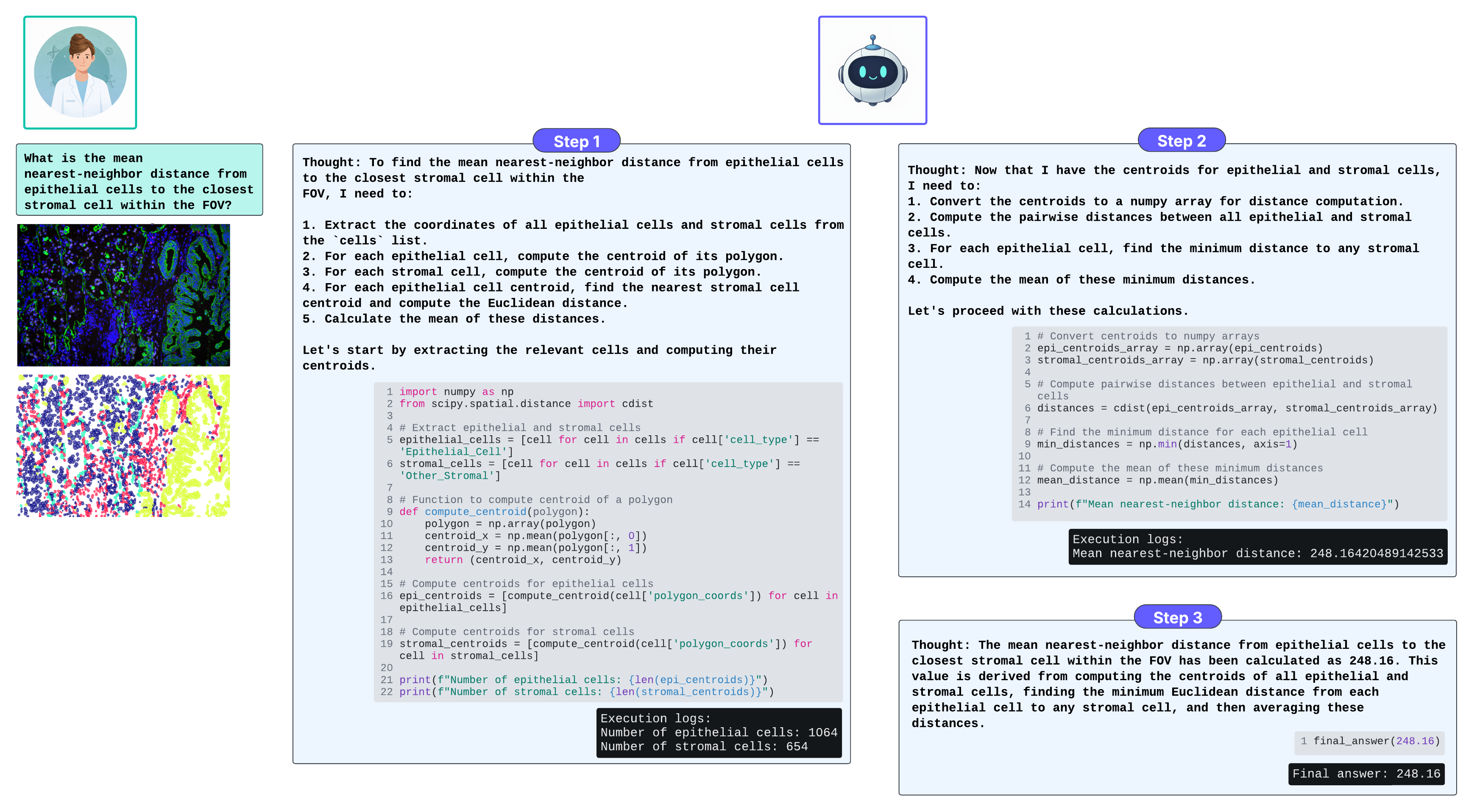}
    \caption{\textbf{Example CodeCytos agent run on an NSCLC tissue field of view (FOV) answering a bioscientist’s query: “What is the mean nearest-neighbor distance from epithelial cells to the closest stromal cell within the FOV?”} The bioscientist provides exactly the question shown in the figure, with no additional task specifications or spatial-analysis instructions. CodeCytos leverages cdist, a highly efficient function within the scipy.spatial.distance module in Python, used to compute the distance between every pair of vectors from two different input matrices (arrays). It acts as a vectorized alternative to calculating distances using nested loops, making it much faster for large datasets.}
    \label{fig:chat_example_2}
\end{figure}

\begin{figure}
    \centering
    \includegraphics[width=0.65\linewidth]{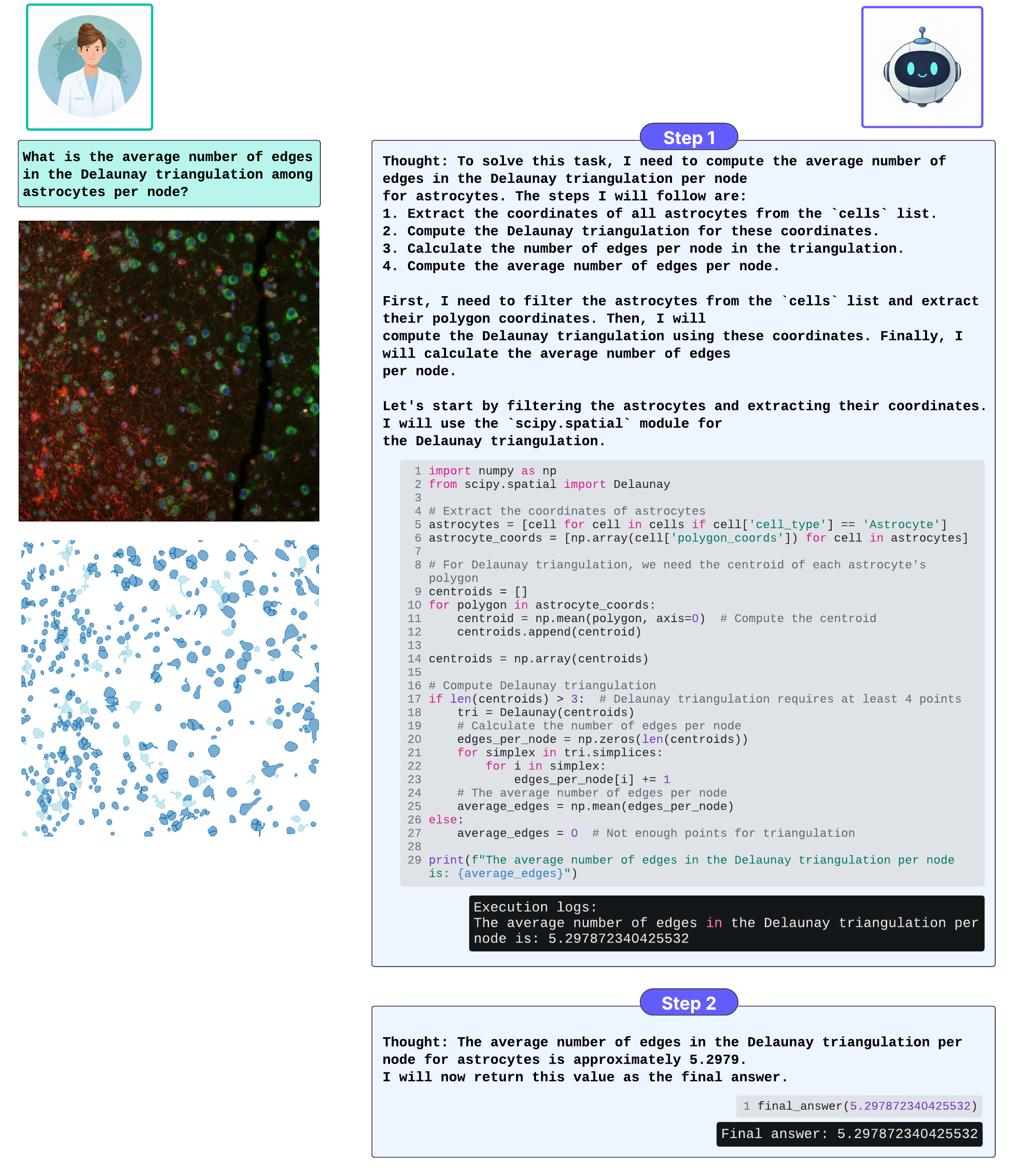}
    \caption{\textbf{Example CodeCytos agent run on an Frontal Cortex tissue field of view (FOV) answering a bioscientist’s query: “What is the average number of edges per node in the Delaunay triangulation among astrocytes?”} The bioscientist provides exactly the question shown in the figure, with no additional task specifications or spatial-analysis instructions.}
    \label{fig:chat_example_3}
\end{figure}

\begin{figure}
    \centering
    \includegraphics[width=\linewidth]{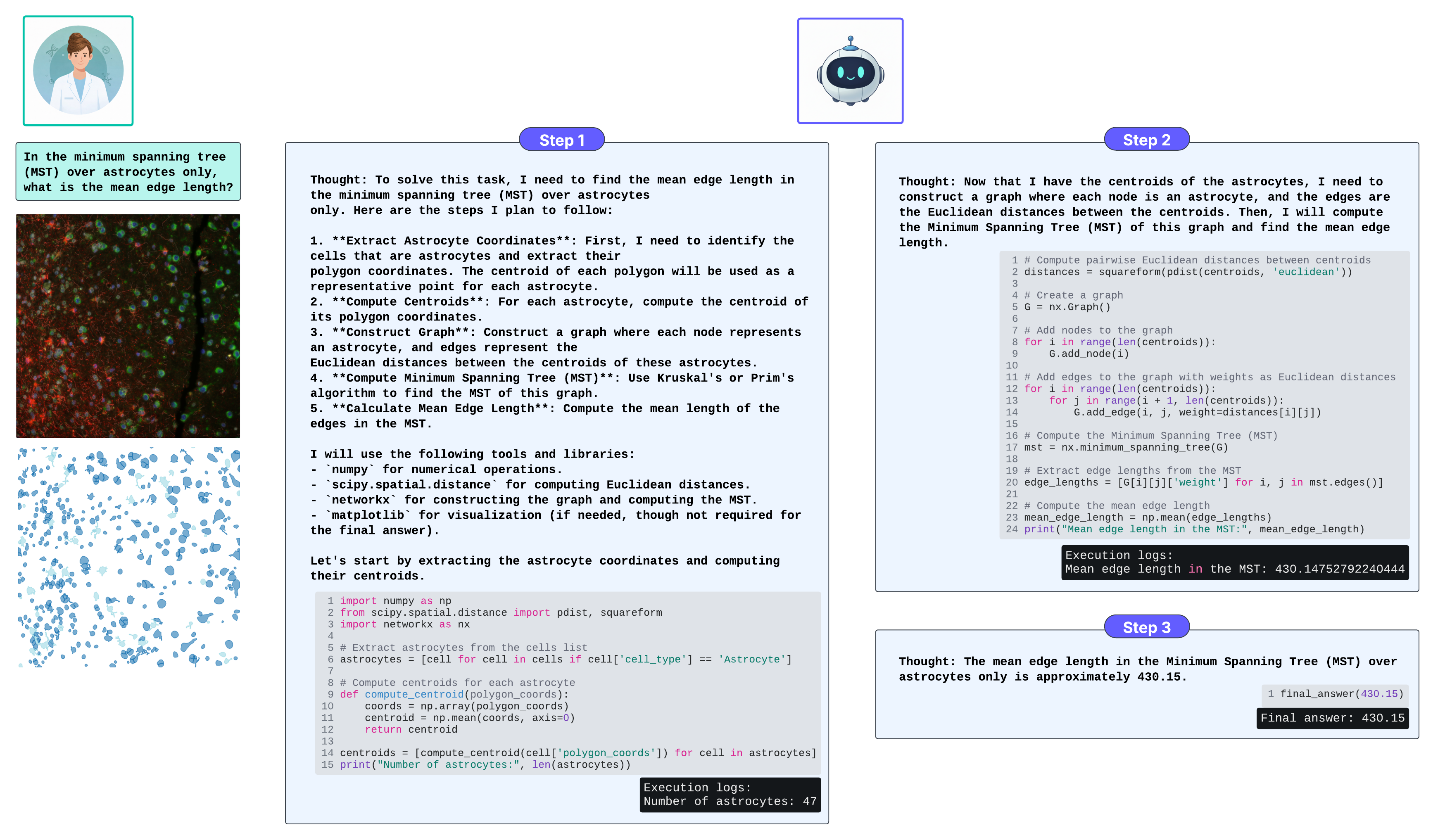}
    \caption{\textbf{Example CodeCytos agent run on an Frontal Cortex tissue field of view (FOV) answering a bioscientist’s query: “In the minimum spanning tree (MST) over astrocytes only, what is the mean edge length?”} The bioscientist provides exactly the question shown in the figure, with no additional task specifications or spatial-analysis instructions. CodeCytos leverages \textit{networkx} to build a graph and compute the minimum spanning tree.}
    \label{fig:chat_example_4}
\end{figure}

\begin{figure}
    \centering
    \includegraphics[width=\linewidth]{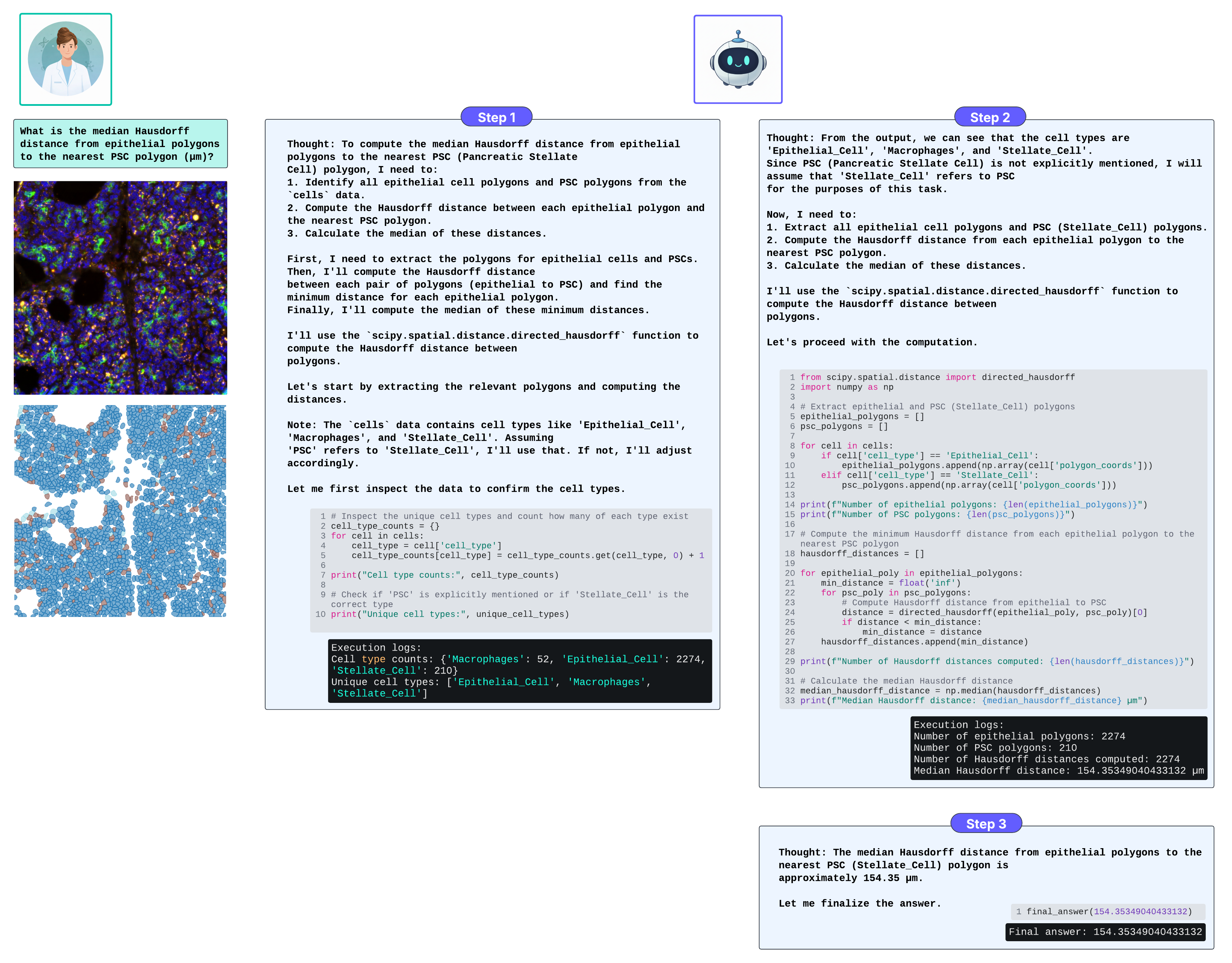}
    \caption{\textbf{Example CodeCytos agent run on an Pancreas tissue field of view (FOV) answering a bioscientist’s query: “What is the median Hausdorff distance from epithelial polygons to the nearest PSC polygon (µm)?”} The bioscientist provides exactly the question shown in the figure, with no additional task specifications or spatial-analysis instructions. CodeCytos interprets “PSC” as pancreatic stellate cells and observes that the classification output contains three cell types: epithelial cells, macrophages, and stellate cells. It therefore assumes that “PSC” corresponds to the stellate-cell class.}
    \label{fig:chat_example_5}
\end{figure}

\begin{figure}
    \centering
    \includegraphics[width=\linewidth]{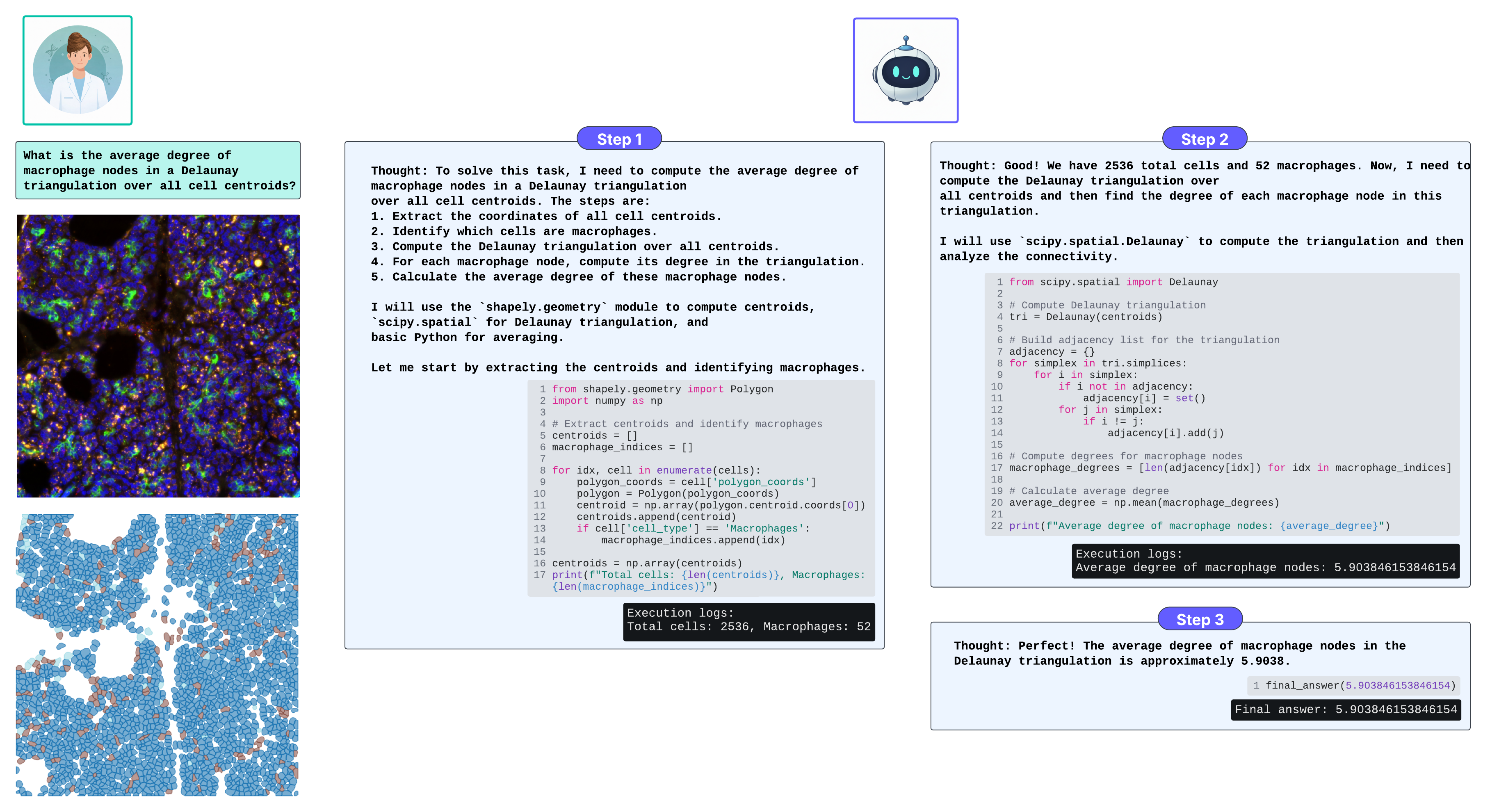}
    \caption{\textbf{Example CodeCytos agent run on an Pancreas tissue field of view (FOV) answering a bioscientist’s query: “What is the average degree of macrophage nodes in a Delaunay triangulation over all cell centroids?”} The bioscientist provides exactly the question shown in the figure, with no additional task specifications or spatial-analysis instructions.}
    \label{fig:chat_example_6}
\end{figure}

\begin{figure}
    \centering
    \includegraphics[width=\linewidth]{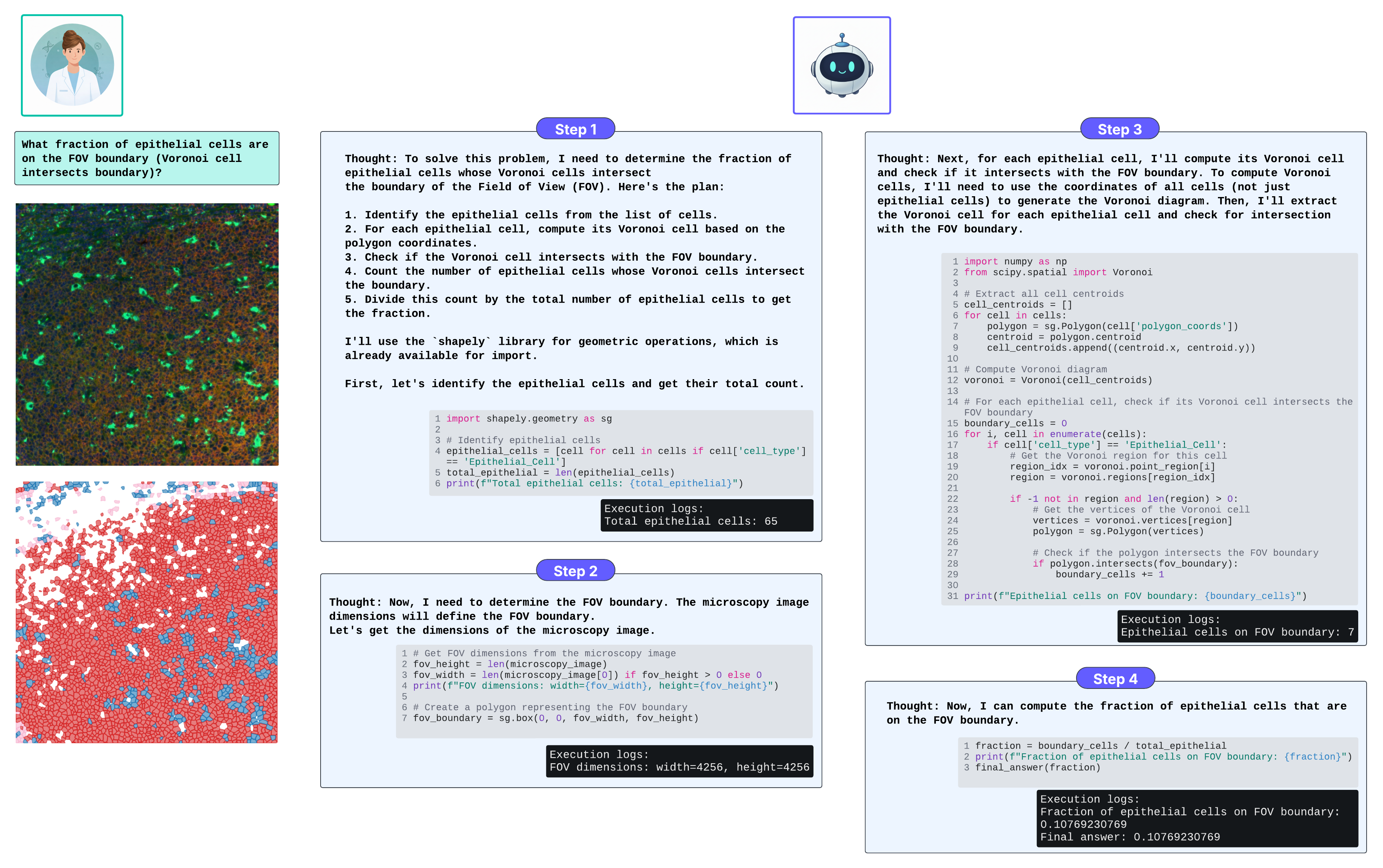}
    \caption{\textbf{Example CodeCytos agent run on an Tonsil tissue field of view (FOV) answering a bioscientist’s query: “What fraction of epithelial cells are on the FOV boundary (i.e., their Voronoi cell intersects the boundary)?”} The bioscientist provides exactly the question shown in the figure, with no additional task specifications or spatial-analysis instructions.}
    \label{fig:chat_example_7}
\end{figure}

\begin{figure}
    \centering
    \includegraphics[width=\linewidth]{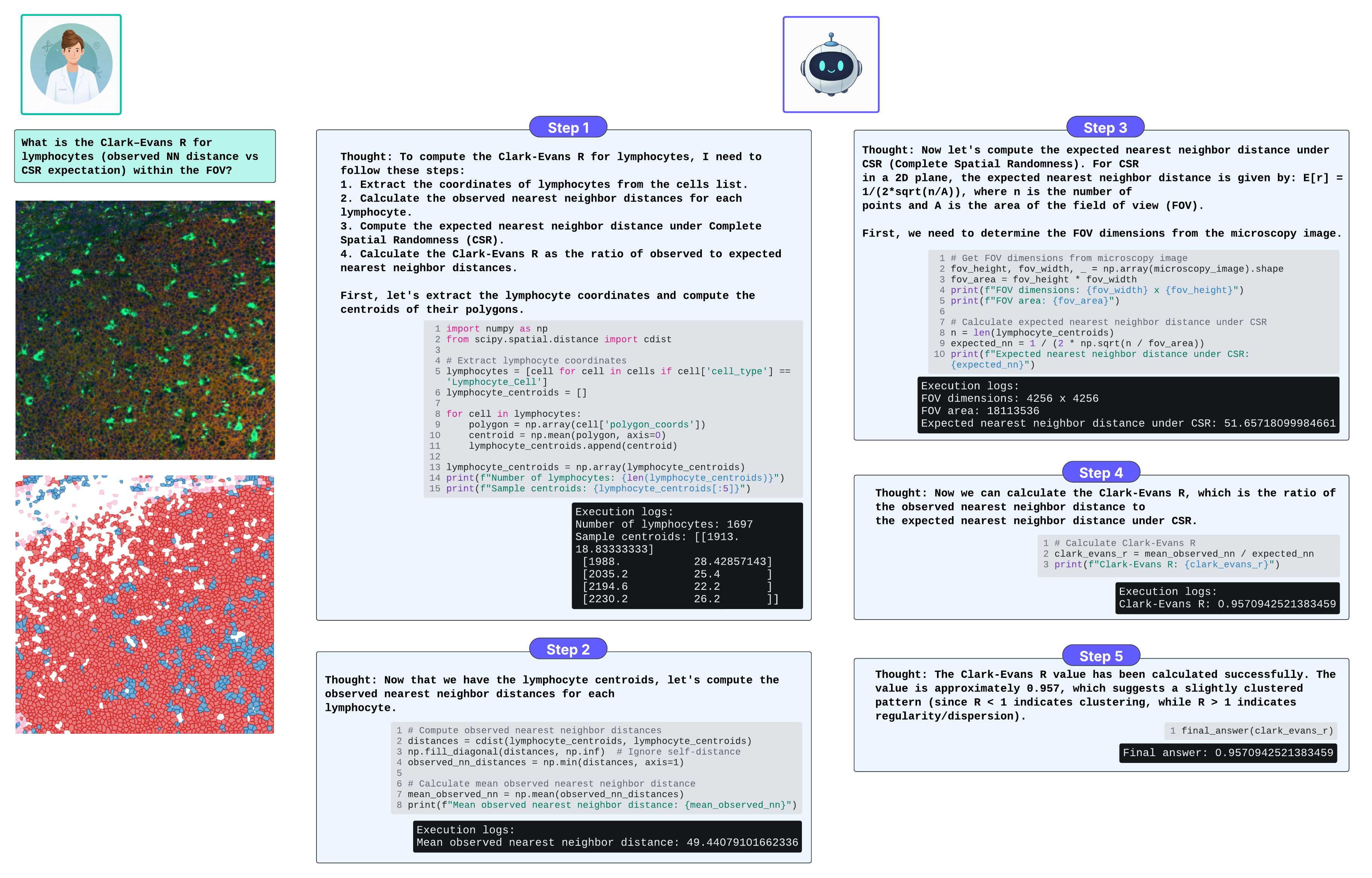}
    \caption{\textbf{Example CodeCytos agent run on an Tonsil tissue field of view (FOV) answering a bioscientist’s query: “What is the Clark–Evans \(R\) for lymphocytes (observed nearest-neighbor distance vs. CSR expectation) within the FOV?”} The bioscientist provides exactly the question shown in the figure, with no additional task specifications or spatial-analysis instructions. CodeCytos recognizes domain-specific terms such as Clark-Evans R or CSR - Complete Spatial Randomness.}
    \label{fig:chat_example_8}
\end{figure}

Figures \ref{fig:chat_example_1}, \ref{fig:chat_example_2}, \ref{fig:chat_example_3}, \ref{fig:chat_example_4}, \ref{fig:chat_example_5}, \ref{fig:chat_example_6}, \ref{fig:chat_example_7}, \ref{fig:chat_example_8}, present several real-world examples of our CodeCytos agent interacting with—and assisting—bioscientists. In each example, the bioscientist poses a specific question about a spatial analysis task, and the agent responds by formulating a step-by-step plan and executing it iteratively. At each step, the agent: (1) writes its reasoning, (2) generates the corresponding commented Python code, and (3) calls an external Python interpreter to execute the code. After execution, the agent inspects the returned outputs (e.g., results or errors) and determines how to proceed. This loop repeats until the task is completed or a predefined maximum number of steps is reached, with each step following the same three-stage structure: reasoning, code generation, and code execution.

It is noteworthy that, even without additional context or task-specific instructions, the agent can still infer what to do and which steps to take given only a small set of essential inputs: the bioscientist’s question, the input cellular image, and the corresponding cell segmentation and classification results. This behavior suggests that the agent can autonomously formulate and execute a reasonable analysis procedure given limited instructions about the spatial analysis task.

We attribute this capability largely to the pretrained knowledge embedded in the LLM backbones, which our agent leverages via the ReAct-based agent design pattern. A representative example appears in Step 2 of our first case study (Fig. \ref{fig:chat_example_1}). When asked, “What is the variance-to-mean ratio (VMR) of T-cell counts in \(30\,\mu\text{m}\) square bins?”, the agent explicitly notes that the physical scale (i.e., \(\mu\)m per pixel) is not provided and therefore assumes \(1\) pixel \(=\) \(1\,\mu\text{m}\). This suggests that the underlying LLMs encode general knowledge about microscopy imaging (e.g., the need for a pixel-to-micron conversion), and that the agent transparently makes such assumptions explicit so users can either provisionally accept them or provide the correct scale.

\subsection*{Agent Evaluation Results on $M^3$ToolEval dataset}

\begin{figure}[h!]
    \centering
    
    \begin{subfigure}{\textwidth}
        \centering
        \rlap{\raisebox{\dimexpr\height-0.8\baselineskip\relax}{\hspace{0.8\baselineskip}\textbf{a}}}%
        \adjustbox{max width=\textwidth}{%
        \begin{tabular}{lcccccccc}
        \toprule
         & \multicolumn{4}{c}{Success Rate (\%, $\uparrow$)} & \multicolumn{4}{c}{Avg. Turns ($\downarrow$)} \\
        action\_mode & CodeAct (zero-shot) & CodeAct-fs (Ours) & JSON & Text & CodeAct & CodeAct-fs (Ours) & JSON & Text \\
        model\_name &  &  &  &  &  &  &  &  \\
        \midrule
        \texttt{NousResearch/Hermes-4-70B} & $\mathbf{51.20}$ & \underline{$45.10$} & $42.70$ & $29.30$ & $\mathbf{6.80}$ & \underline{$7.10$} & $8.10$ & $8.50$ \\
        \texttt{facebook/cwm} & $37.80$ & $\mathbf{54.90}$ & $40.20$ & \underline{$43.90$} & $8.70$ & $\mathbf{7.90}$ & $8.50$ & \underline{$8.40$} \\
        \texttt{mistralai/Devstral-2-123B-Instruct-2512} & \underline{$63.40$} & $\mathbf{67.10}$ & $57.30$ & $56.10$ & $\mathbf{6.10}$ & \underline{$6.60$} & $7.60$ & $7.60$ \\
        \texttt{mistralai/Devstral-Small-2-24B-Instruct-2512} & \underline{$64.60$} & $\mathbf{68.30}$ & $50.00$ & $46.30$ & \underline{$6.50$} & $\mathbf{6.20}$ & $7.80$ & $8.20$ \\
        \texttt{mistralai/Devstral-Small-2505} & \underline{$54.90$} & $\mathbf{58.50}$ & $48.80$ & $46.30$ & $\mathbf{6.80}$ & \underline{$6.90$} & $7.90$ & $8.10$ \\
        \texttt{moonshotai/Kimi-Linear-48B-A3B-Instruct} & \underline{$35.40$} & $\mathbf{48.80}$ & $28.00$ & $25.60$ & \underline{$8.50$} & $\mathbf{7.60}$ & $8.80$ & $8.80$ \\
        \texttt{qwen/Qwen2.5-Coder-32B-Instruct} & $\mathbf{48.80}$ & $43.90$ & \underline{$47.60$} & $39.00$ & $\mathbf{6.80}$ & \underline{$6.90$} & $8.00$ & $8.30$ \\
        \texttt{qwen/Qwen3-Coder-30B-A3B-Instruct} & \underline{$59.80$} & $\mathbf{63.40}$ & $45.10$ & $29.30$ & \underline{$7.20$} & $\mathbf{7.10}$ & $8.20$ & $8.70$ \\
        \texttt{XiaomiMiMo/MiMo-V2-Flash} & $39.00$ & $\mathbf{61.00}$ & \underline{$52.40$} & $47.60$ & $8.10$ & $\mathbf{7.10}$ & \underline{$7.90$} & $8.10$ \\
        \texttt{zai-org/GLM-4.5-Air} & $\mathbf{46.30}$ & $41.50$ & \underline{$42.70$} & \underline{$42.70$} & \underline{$8.40$} & \underline{$8.40$} & \underline{$8.40$} & $\mathbf{8.20}$ \\
        \bottomrule
        \end{tabular}%
        }
        \label{subfig:m3-eval-table}
    \end{subfigure}
    
    \vspace{1em} 
    
    \begin{subfigure}{\textwidth}
        \centering
        \rlap{\raisebox{\dimexpr\height-0.8\baselineskip\relax}{\hspace{0.8\baselineskip}\textbf{b}}}%
        \includegraphics[width=\linewidth]{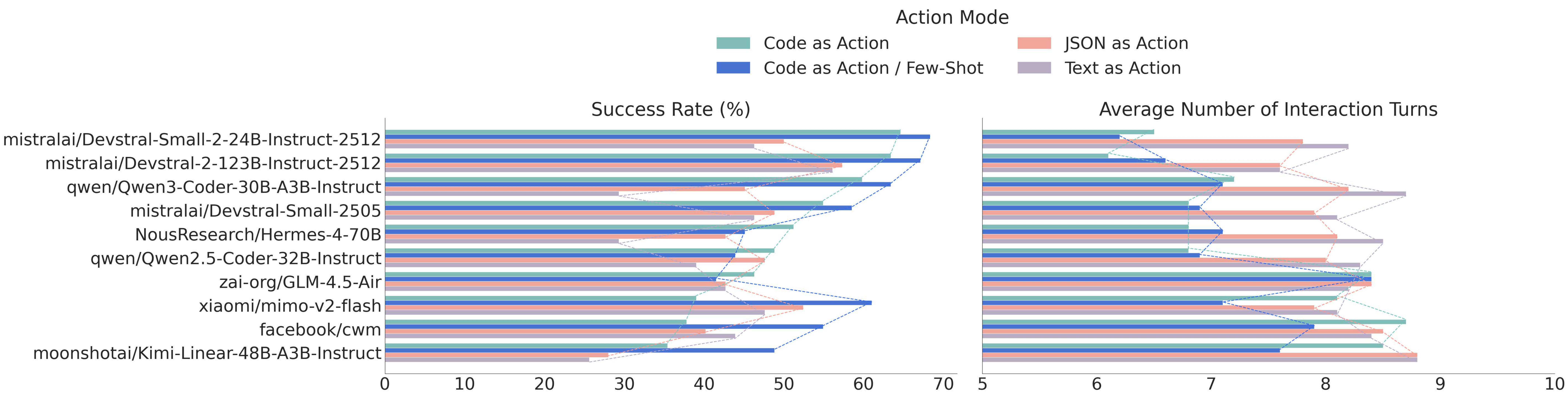}        
        \label{subfig:m3-eval-plot}
    \end{subfigure}
    
    \caption{\textbf{Effect of domain-agnostic few-shot coding--reasoning demonstrations on M$^3$ToolEval.} We compare \textsc{CodeAct} (zero-shot) with \textsc{CodeAct-fs} (ours; \textsc{CodeAct} augmented with domain-agnostic few-shot demonstrations) against JSON and Text action modes on the original M$^3$ToolEval benchmark from CodeAct~\cite{wang2024executable} (82 human-curated, multi-turn, multi-tool tasks). \textbf{a,} Per-model success rate (\%, higher is better) and average number of dialogue turns (lower is better); best and second-best results are shown in \textbf{bold} and \underline{underline}, respectively. \textbf{b,} Aggregate performance comparison across models on M$^3$ToolEval. Compared to CodeCytos on our spatial cellular benchmark, few-shot prompting yields smaller (but still meaningful) gains on \(M^3\)ToolEval across backbones, likely because 
    \(M^3\)ToolEval mainly rewards correct multi-turn tool selection/invocation (often achievable with simple text/JSON actions) rather than complex coding reasoning, leaving less headroom for improvement from coding--reasoning exemplars.
}
    
    \label{fig:combined_m3tooleval_analysis}
\end{figure}

\subsubsection*{Evaluation Results}
Figure \ref{fig:combined_m3tooleval_analysis} shows evaluation results on the \(M^3\)ToolEval dataset (the original CodeAct benchmark \cite{wang2024executable}) under four settings: (1) CodeAct as action, (2) CodeAct as action (few-shot), (3) JSON as action, and (4) Text as action. We report two metrics used in the original CodeAct paper: success rate and average turns. Success rate is the percentage of model-generated answers that match the ground-truth solutions, while average turns is the mean number of turns across all evaluated instances.

Adding the same set of few-shot examples as in our previous experiment, i.e., without crafting any domain-specific in-context examples from the \(M^3\)ToolEval training set, still improves the CodeAct agent’s performance in both success rate and interaction turns. Across multiple LLM backbones with strong coding capabilities, the CodeAct few-shot setting most frequently ranks first, followed by the original zero-shot CodeAct setting. The JSON and Text action formats also achieve competitive performance, but they consistently lag behind the CodeAct and CodeAct-few-shot settings.

\subsubsection*{Observations and Explanations}
We observe that for most coding LLM backbones, adding a few random few-shot coding-reasoning exemplars significantly boosts performance on the $M^3$ToolEval dataset. For example, on Facebook-CWM, performance increases from 37.8 to 54.9 when few-shot exemplars are added to the original CodeAct. Similar observations hold for Kimi-Linear-48B-A3B-Instruct and MiMo-V2-Flash. For Devstral-2-123B-Instruct-2512, Devstral-Small-2-24B-Instruct-2512, and Devstral-Small-2505, performance improves more modestly. The highest result is obtained with Devstral-Small-2-24B-Instruct-2512 under the CodeAct-fewshot setting. Overall, task-agnostic few-shot coding–reasoning examples can significantly improve performance without costly training, surpassing the best open-source result reported in the original paper (13.4\%). We attribute the improvement to two factors: (1) a stronger core LLM backbone, and (2) the inclusion of random few-shot exemplars that better guide the model’s behavior.

In some instances, incorporating few-shot exemplars degrades performance, as observed for Hermes-4-70B, Qwen2.5-Coder-32B-Instruct, and GLM-4.5-Air \ref{fig:combined_m3tooleval_analysis}. Moreover, compared with our earlier CodeCytos experiments on a spatial cellular analysis task, the improvements from few-shot prompting on \(M^3\)ToolEval—although meaningful—are consistently less pronounced across backbones. A plausible explanation is that \(M^3\)ToolEval is a multi-turn, multi-tool benchmark that primarily rewards correct tool selection and invocation, rather than the more complex programming and reasoning emphasized in our spatial cellular analysis benchmark. Consistent with this interpretation, strong performance on \(M^3\)ToolEval can already be achieved with simple action formats (e.g., plain text or JSON) \ref{fig:combined_m3tooleval_analysis}, leaving less headroom for additional improvements from coding few-shot exemplars.

\section*{Discussion}

In this paper, we present CodeCytos, an intelligent CodeAct agent that supports scientists in spatial molecular imaging analysis, improving the efficiency of biomarker discovery. This work is motivated by limitations of conventional analysis software, which typically provides only a fixed set of predefined functions. When bioscientists require an unsupported analysis or need to extract spatial cellular features not available in existing tools, they must either switch to alternative software—if such software exists—or seek assistance from computational scientists or software engineers. These workarounds are often time-consuming and disruptive, and can slow the overall discovery process. CodeCytos addresses this gap by interpreting bioscientists’ natural-language requests and automatically generating and executing code to extract the requested spatial cellular features, thereby streamlining biomarker discovery.

To evaluate the effectiveness of CodeCytos, we compiled an expert-curated set of questions targeting diverse cellular spatial features across four datasets representing distinct tissue types. We evaluate in a minimal-prompt regime: questions are presented without task instructions or additional contextual hints about the underlying spatial analysis, minimizing prompt-engineering effects and testing whether a method can infer the required analysis and execute it reliably. The questions span multiple spatial feature categories, including neighbor-based features, graph-based features, and spatial statistics. In total, the benchmark includes 548 questions, each paired with 20 fields of view (FOVs), yielding 10,960 image–question pairs.

We evaluate the performance of CodeCytos using multiple metrics to characterize its effectiveness across different settings. The success rate quantifies the agent's performance on a single run/attempt. We also report pass@k for \(k \in \{5, 10, 20\}\) to measure performance when multiple runs/attempts are carried out. Finally, we propose a summary metric that aggregates pass@k over all \(k\) values considered in this study (from 1 to 20 attempts), which we term the Area Under the Pass@k Curve (AUP@k).

Our CodeCytos Agent, which employs a straightforward yet effective prompting strategy utilizing few-shot guided coding-reasoning examples, consistently outperforms other baseline methods across various datasets and tasks. However, when utilizing different coding LLM backbones, the addition of few-shot coding-reasoning guided examples yields variable improvements depending on the dataset and task. This indicates that future research could explore more effective approaches to achieve consistent performance enhancements across different Coding LLM backbones.

Overall, our work is driven by the goal of connecting bioscientist experts with a new class of software that leverages coding agents, offering greater flexibility than traditional tools for transforming and streamlining biomarker discovery.

It is still worth noting that CodeCytos is not intended to replace conventional spatial cellular analysis software or packages. Rather, it enables scientists to query specific spatial cellular features that are not supported by existing tools, offering a practical alternative to requesting custom implementations from computational scientists or software engineers. Moreover, it also complements conventional softwares by leveraging existing functionality when appropriate, while generating additional code when needed. In this way, CodeCytos is designed to bridge gaps in current analysis workflows and facilitate biomarker discovery for bioscientists.

\section*{Methods}

\subsection*{Dataset Collection and Expert Curation}
For cellular tissue images, we leverage publicly available datasets generated with the CosMx Spatial Molecular Imager (SMI) \cite{nanostring_website} and released by the manufacturer. In this study, we use four datasets representing distinct human tissue types: Frontal Cortex, Non-Small Cell Lung Cancer, Pancreas, and Tonsil.

We first define a broad panel of algorithmic, spatially informed cellular features that represent core analyses in spatial molecular imaging datasets, including nearest-neighbor distances, local neighborhood composition, geometric descriptors, spatial statistics, and graph-based relational measures. Using this panel as the specification, we leverage an LLM to express each feature as a concise natural-language question presented without task instructions or additional context, thereby enforcing a strict minimal-prompt evaluation setting. We then manually audit the generated questions to confirm a faithful, one-to-one correspondence with the intended features.

All tissue-specific question sets are subsequently curated and validated by an expert biologist to ensure biological relevance, clarity, and interpretability. We incorporate this feedback in a second collection round, expanding coverage by adding more diverse question types, followed by an additional validation cycle. After consensus in the second review, we merge questions across iterations to form the final benchmark, yielding over 100 expert-validated questions for each of the four tissue types.

To obtain ground-truth answers for these questions, we simulate an analysis workflow, similar to Vaidya et al. \cite{vaidya2025nova}, in which Machine-Learning (ML) engineers will provide human-written code when being provided with (i) a tissue field-of-view (FOV) and (ii) a biologist-authored question specifying a target category of spatial features. For each image–question pair, each ML engineer (1) writes code to load and preprocess the image, (2) applies appropriate tools for cell segmentation and cell type classification, and (3) implements the computation required for the requested feature category (e.g., nearest-neighbor distances, neighborhood composition, geometric metrics, or graph-based descriptors) using the cell boundaries and cell types predictions. The resulting code is executed to produce the final quantitative output, which we treat as the ground-truth answer for validation of our agent system.

For each tissue type, we include 20 different Field-Of-View images, each paired with a curated set of over 100 questions spanning multiple spatial feature categories. In total, the constructed dataset contains 10,960 image–question pairs with associated ground-truth outputs.

\subsection*{CodeCytos Agent Architecture and Components}
\subsubsection*{Problem Setup}
We formulate cellular image analysis as a sequential decision-making problem in which an agent interacts with an environment to complete a user-specified task. At time step \(t\), the agent receives an observation \(o_t \in O\) (e.g., user instructions, uploaded images, intermediate outputs, and tool responses) and selects an action \(a_t \in A\) according to a policy \(\pi(a_t \mid c_t)\), where the agent context is the interaction history

\begin{equation}
    c_t = (o_1, a_1, \ldots, o_{t-1}, a_{t-1}, o_t)
\end{equation}

The goal is to iteratively update the context and execute actions until the task is solved.

\subsubsection*{ReAct-Style Interleaving of Reasoning and Acting}
To enable multi-step problem solving, we adopt the ReAct principle~\cite{yao2022react} by augmenting the agent's action space with a reasoning channel. Specifically, we consider an augmented action space \(\tilde{A} = A \cup L\), where \(L\) denotes language ``thought'' actions that do not directly affect the environment. A thought action \(\tilde{a}_t \in L\) is used to summarize the current context, identify missing information, and plan the next tool call(s). Since thought does not execute in the environment, it produces no external observation; instead, it updates the internal context:

\begin{equation}
    c_{t+1} = (c_t, \tilde{a}_t).    
\end{equation}

We then alternate between generating (i) a thought that structures the next step and (ii) an environment-facing action that performs computation, producing a new observation that is appended to the context. This yields iterative trajectories of the form thought--action--observation repeated over multiple steps.

\subsubsection*{CodeAct-Style Executable Actions}
While ReAct typically uses free-form tool calls and task-specific prompt demonstrations, we implement the \emph{act} component with CodeAct \cite{wang2024executable}, in which environment-facing actions are emitted as executable Python code. Concretely, CodeCytos uses code as a unified action representation that (i) invokes analysis tools, (ii) expresses data and control flow, and (iii) stores intermediate results in variables for reuse across steps. This design reduces reliance on handcrafted in-context demonstrations and allows the agent to compose multiple operations within a single code action when appropriate, yielding strong performance even in a zero-shot setting \cite{wang2024executable}.

\subsubsection*{Multi-Turn Execution with Observation-Driven Self-Correction}
Each code action is executed in a sandboxed environment, and the resulting outputs---including tool returns, logs, and error messages---are captured as observations. We denote these execution results as \(o_{t+1}\) and append them to the running context. If execution fails, the error message provides structured feedback for the next step: the agent revises the code accordingly (self-debugging) and re-executes until the action succeeds or the interaction reaches the maximum step budget.

\subsubsection*{CodeCytos Tool Interface for Cellular Image Analysis}
CodeCytos is instantiated with two primary tool categories: (1) \emph{cell segmentation} and (2) \emph{cell typing/classification}. Because different datasets and imaging modalities may favor different methods, a pool of cell segmentation \cite{stringer2021cellpose, pachitariu2022cellpose, pachitariu2025cellpose, goldsborough2024instanseg, stevens2022stardist, archit2025segment, van2016deep} and cell classification \cite{van2016deep, ghahremani2022deepliif, horst2024cellvit, satija2015spatial, dries2021giotto, biancalani2021deep, kleshchevnikov2022cell2location} tools can be integrated, allowing the bioscientist to select which ones to use for a given study. At runtime, CodeCytos conditions on the user's selection and calls the corresponding tool functions in code, applying them to the uploaded cellular image(s). Beyond these domain tools, the agent may import and use general-purpose Python packages as needed for preprocessing, quality control, feature extraction, result aggregation, and visualization, enabling end-to-end analysis within the same iterative thought--code--observation loop.

\subsubsection*{Overall Algorithm}
Across a task, CodeCytos repeatedly:
\begin{enumerate}
    \item \textbf{Thought (ReAct):} reason over \(c_t\), plan the next step, and decide which tool(s) and parameters are needed.
    \item \textbf{Act (CodeAct):} generate executable Python code that calls the selected segmentation/classification tool(s) and any auxiliary packages.
    \item \textbf{Observe:} execute the code to obtain outputs or errors, forming \(o_{t+1}\) and updating the context.
\end{enumerate}
This integration of ReAct \cite{yao2022react} (explicit iterative reasoning) and CodeAct \cite{wang2024executable}  (code-based acting with execution feedback) provides a practical methodology for constructing CodeCytos as a multi-step, tool-using agent for cellular spatial image analysis.

\subsection*{CodeCytos Tooling via Documented Function Interfaces}
We implement two core tool functions—one for cell segmentation and one for cell classification—and expose them to the CodeCytos agent as callable tools. Each tool is implemented as a Python function with a detailed docstring that specifies: (i) the tool's purpose and the task(s) it supports, (ii) input parameter metadata (e.g., name, type, and semantic role), and (iii) return variable metadata (e.g., type and meaning of outputs). We incorporate these docstrings into the agent's available context as part of the environment specification, enabling CodeCytos to condition its code generation on explicit tool-interface documentation when selecting tools, constructing function calls, and interpreting returned outputs.

\subsection*{LLM Backbone Selection and Inference Serving}
To instantiate CodeCytos, we select large language models (LLMs) with strong code-generation capability, since CodeCytos emits executable Python code as its primary action representation. The coding LLMs used in our experiments include: Qwen2.5-Coder-32B-Instruct \cite{hui2024qwen2}, Qwen3-Coder-30B-A3B-Instruct \cite{yang2025qwen3}, Facebook-Code World Model (CWM) \cite{copet2025cwm}, Devstral-Small-2505 \cite{rastogi2025devstral}, Devstral-2-123-Instruct-2512 \cite{rastogi2025devstral}, Kimi-Linear-48B-A3B-Instruct \cite{team2025kimi}, and GLM-4.5-Air \cite{zeng2025glm}. These models span a range of scales, from approximately 24 billion to 123 billion parameters. For high-throughput inference, we deploy the models using either vLLM \cite{kwon2023efficient} or SGLang \cite{zheng2024sglang} as the inference serving backend.

\subsection*{Agent and Language Model Configurations}
For all CodeCytos experiments, we set the maximum agent trajectory length to \(10\) steps. Each run terminates early upon successful task completion; otherwise, it stops when the step limit is reached. Across all coding LLM backbones used in CodeCytos, we use a sampling temperature of \(0.7\) and cap generation at \(2048\) tokens per step.

\subsection*{Few-shot Coding-Reasoning Prompting for CodeCytos}
We use few-shot in-context prompting to improve the reliability of \emph{coding-reasoning} behaviors in our CodeAct-based CodeCytos agent. Let \(x\) denote a task instance (e.g., image(s), user request, and any auxiliary metadata), and let the agent interact with an environment over discrete steps \(t=1,\ldots,T\). At step \(t\), the agent conditions on the context
\[
c_t = (o_1,a_1,\ldots,o_{t-1},a_{t-1},o_t),
\]
and samples the next action \(a_t\) from a backbone LLM \(p_\theta\) as
\[
a_t \sim p_\theta(\cdot \mid c_t).
\]
In CodeCytos, actions are emitted in a \emph{structured} format consisting of (i) a reasoning segment and (ii) an executable code segment:
\[
a_t \equiv \big(r_t,\, z_t\big),
\]
where \(r_t\) is a natural-language reasoning trace (enclosed by \texttt{<thought>} tags) and \(z_t\) is Python code (enclosed by \texttt{<code>} tags). The environment executes \(z_t\) and returns an observation \(o_{t+1}\) (e.g., tool outputs, logs, or exceptions), which updates the context.

\paragraph{Background and motivation.}
In ReAct~\cite{yao2022react}, tool actions are expressed in the language space and are typically handcrafted as textual tool calls. As a result, few-shot demonstrations are required to teach models the desired \emph{thought--action--observation} structure and domain-specific action semantics. ReAct constructs multi-step annotated trajectories from training data (e.g., HotPotQA \cite{yang2018hotpotqa}, FEVER \cite{thorne2018fever}, and ALFWorld \cite{shridhar2020alfworld}) and prepends them as exemplars for in-context learning. Although effective, creating such demonstrations can require substantial prompt engineering and manual annotation, and scaling trace supervision can be costly.

CodeAct~\cite{wang2024executable} reduces this burden by consolidating actions into a unified, executable Python action space. This leverages the coding priors of LLMs and can work well in zero-shot settings~\cite{wang2024executable}. However, prior results also indicate a substantial performance gap between open-source and proprietary backbones in the zero-shot regime.

\paragraph{Few-shot prompting as behavior specification.}
To improve robustness, particularly for open-source coding LLMs, we prepend a small set of \emph{general} coding-reasoning exemplars \(D_{\mathrm{fs}}=\{e_i\}_{i=1}^{m}\) to the system prompt. Each exemplar \(e_i\) consists of a short multi-step interaction that demonstrates:
\begin{itemize}
    \item the expected reasoning style \(r_t\),
    \item executable code generation \(z_t\),
    \item adherence to the output template (e.g., \texttt{<thought>} and \texttt{<code>} tags),
    \item and next-step planning or correction based on execution observation \(o_{t+1}\).
\end{itemize}
Importantly, these exemplars are \emph{not domain-specific} and are intentionally unrelated to spatial cellular analysis; they are used purely to specify the agent's interaction protocol and coding-reasoning behavior.

Formally, few-shot prompting conditions the model on a demonstration set \(D_{\mathrm{fs}}\), thereby modifying the conditional distribution over actions:
\[
a_t \sim p_\theta(\cdot \mid D_{\mathrm{fs}}, c_t).
\]
In our setting, \(D_{\mathrm{fs}}\) guides the LLM's coding-and-reasoning behavior and potentially reduces template-format violations through given examples by reinforcing the required structure: the agent's thought must appear within \texttt{<thought></thought>}, code within \texttt{<code></code>}, and observations within \texttt{<obs></obs>}.


\subsection*{Baseline Comparisons}
\subsubsection*{Spatial Cellular Tissue Dataset}
For the main experiments on spatial cellular images, we compare two overarching settings: (i) a \emph{tool-augmented LLM} baseline and (ii) our proposed \emph{CodeAct-based CodeCytos} agent. Here, \emph{tool-augmented LLM} refers to an LLM that can access external tools and files, including a Python interpreter for code execution, the uploaded image(s), and intermediate outputs such as cell segmentation and classification results.

\noindent \textit{Tool-augmented LLM baselines.}
We evaluate three variants:
\begin{enumerate}
    \item Tool-augmented LLM,
    \item Tool-augmented LLM with zero-shot Chain-of-Thought (CoT),
    \item Tool-augmented LLM with few-shot Chain-of-Thought (CoT).
\end{enumerate}

\noindent \textit{CodeCytos (CodeAct-based) variants.}
We evaluate three variants:
\begin{enumerate}
    \item CodeAct-based CodeCytos (Original CodeAct -- zero-shot),
    \item CodeAct-based CodeCytos with one-shot coding-reasoning prompting,
    \item CodeAct-based CodeCytos with few-shot coding-reasoning prompting.
\end{enumerate}

\noindent Few-shot exemplar construction:
Across all settings, the one-shot and few-shot exemplars are \emph{not domain-specific}: they are general coding-reasoning trajectories with tool calls, and are not curated from the spatial cellular dataset. 

\subsubsection*{$M^3$ToolEval Dataset}

For our ablation study on the original CodeAct benchmark, \(M^3\)ToolEval, we adopt the same three baseline settings as in the paper: (1) Text-as-Action, (2) JSON-as-Action, and (3) Code-as-Action. We then augment the Code-as-Action setting with a small set of few-shot coding--reasoning exemplars to assess whether they further improve performance. These exemplars are identical to those used in our proposed Spatial Cellular Analysis benchmark.

\subsection*{Evaluation and Performance Metrics}
To evaluate CodeCytos, we report two primary metrics: \emph{success rate} and \emph{pass@}k. We define success rate as the fraction of tasks successfully completed with a correct outcome, i.e., the agent's final answer matches the ground-truth answer. Formally, letting \(N_{\text{success}}\) denote the number of tasks solved correctly and \(N_{\text{total}}\) denote the total number of tasks, the success rate is
\begin{equation}
    \text{Success Rate} = \frac{N_{\text{success}}}{N_{\text{total}}} \times 100\%.
\end{equation}

We additionally report \(\text{pass@}k\) \cite{chen2021evaluating}, which measures the probability that at least one correct solution appears among \(k\) runs/attempts. For a given problem, let \(n\) be the total number of attempts and \(c\) be the number of correct attempts among the \(n\) attempts. We compute \(\text{pass@}k\) as
\begin{equation}
    \text{pass@}k = \mathbb{E}_{\text{Problems}} \left[ 1 - \frac{\binom{n-c}{k}}{\binom{n}{k}} \right].
\end{equation}
In our experiments, we generate up to \(K=20\) agent runs/attempts per problem, and evaluate \(\text{pass@}k\) for \(k \in \{1,\ldots,K\}\).

Since \(\text{pass@}k\) yields a sequence of values across different \(k\), we further summarize performance with a single scalar metric: the area under the \(\text{pass@}k\) curve, denoted as \(\text{AUP@}k\), defined as
\begin{equation}
    \text{AUP@}k = \int_{k=k_{\min}}^{k_{\max}} \text{pass@}k \, dk,
\end{equation}
where \(k_{\min}=1\) and \(k_{\max}=K\) in this work.

\section*{Author contributions}

\textbf{Conceptualization:} Hung Q. Vo, Huy Q. Vo, Son T. Ly, Hien V. Nguyen \\
\textbf{Methodology:} Hung Q. Vo, Huy Q. Vo, Hien V. Nguyen \\
\textbf{Investigation:} Hung Q. Vo, Huy Q. Vo, Anh-Vu Nguyen, Hien V. Nguyen \\
\textbf{Formal analysis:} Hung Q. Vo, Huy Q. Vo, Hien V. Nguyen \\
\textbf{Conceptual framing:} Hung Q. Vo, Huy Q. Vo, Zhihao Wan, Hong Zhao, Jianting Sheng, Stephen T. C. Wong, Hien V. Nguyen \\
\textbf{Resources:} Stephen T. C. Wong, Hien V. Nguyen \\
\textbf{Writing -- original draft:} Hung Q. Vo, Huy Q. Vo \\
\textbf{Writing -- review \& editing:} All authors \\
\textbf{Supervision:} Stephen T. C. Wong, Hien V. Nguyen \\
\textbf{Funding acquisition:} Hong Zhao, Jianting Sheng, Stephen T. C. Wong, Hien V. Nguyen

\section*{Data Availability}
The Spatial Molecular Tissue Images used in this study are available from NanoString at \url{https://nanostring.com/products/cosmx-spatial-molecular-imager/ffpe-dataset/}. The expert-curated benchmark of spatial cellular features for four tissue types (Frontal Cortex, NSCLC, Pancreas, and Tonsil) will be released upon acceptance at \url{https://github.com/hula-ai/CodeCytos}.

\section*{Code Availability}
The CodeCytos source code will be released upon acceptance at \url{https://github.com/hula-ai/CodeCytos}.

\bibliography{sample}










\end{document}